\documentclass{article}

\usepackage{microtype}
\usepackage{graphicx}
\usepackage{subcaption}
\usepackage{booktabs} % for professional tables
\usepackage{soul}

\usepackage{tikz}
\usepackage{amsfonts}
\usepackage{amsmath}
\usepackage{amssymb}
\usepackage{nicefrac}
\usepackage{siunitx}
\usepackage{textcomp}
\usepackage{url}
\usepackage{color}
\usepackage{xcolor,colortbl}
\usepackage{xspace}
\usepackage{times}
\usepackage{textcomp}
\usepackage{lipsum}
\usepackage{pifont}% http://ctan.org/pkg/pifont
\usepackage{multirow}
\usepackage{aliascnt}
\usepackage{threeparttable}

\newtheorem{theorem}{Theorem}

\newcommand{\x}{\mathbf{x}}
\newcommand{\y}{\mathbf{y}}

\newcommand{\cmark}{\ding{51}}%
\newcommand{\xmark}{\ding{55}}%

\usepackage{hyperref}

\usepackage[accepted]{icml2021}

\icmltitlerunning{Kernel Continual Learning}

\begin{document}

\twocolumn[
\icmltitle{Kernel Continual Learning}
%\cs{Title is more catchy, but replay-free is not novel, table 1 is full of replay-free methods. What does the RFF-kernel enable that others cannot do? Focus not on the how, but the what or the why.}\xz{Table 1 is misleading, to the best of our knowledge, we are the only work using memory without replay.}
%\icmltitle{Free of Memory Reply: Kernel Continual Learning with Variational Random Features}
%\icmltitle{Kernel Continual Learning without Experience Replay}

% It is OKAY to include author information, even for blind
% submissions: the style file will automatically remove it for you
% unless you've provided the [accepted] option to the icml2021
% package.

% List of affiliations: The first argument should be a (short)
% identifier you will use later to specify author affiliations
% Academic affiliations should list Department, University, City, Region, Country
% Industry affiliations should list Company, City, Region, Country

% You can specify symbols, otherwise they are numbered in order.
% Ideally, you should not use this facility. Affiliations will be numbered
% in order of appearance and this is the preferred way.
\icmlsetsymbol{equal}{*}

\begin{icmlauthorlist}
\icmlauthor{Mohammad Mahdi Derakhshani}{uva}
\icmlauthor{Xiantong Zhen}{uva,iiai}
\icmlauthor{Ling Shao}{iiai}
\icmlauthor{Cees G.~M. Snoek}{uva}
\end{icmlauthorlist}

\icmlaffiliation{uva}{AIM Lab, University of Amsterdam, The Netherlands}
\icmlaffiliation{iiai}{Inception Institute of Artificial Intelligence, UAE}

\icmlcorrespondingauthor{M. Derakhshani}{m.m.derakhshani@uva.nl}
\icmlcorrespondingauthor{X. Zhen}{x.zhen@uva.nl}

% You may provide any keywords that you
% find helpful for describing your paper; these are used to populate
% the "keywords" metadata in the PDF but will not be shown in the document
\icmlkeywords{Machine Learning, ICML}

\vskip 0.3in
]

% this must go after the closing bracket ] following \twocolumn[ ...

% This command actually creates the footnote in the first column
% listing the affiliations and the copyright notice.
% The command takes one argument, which is text to display at the start of the footnote.
% The \icmlEqualContribution command is standard text for equal contribution.
% Remove it (just {}) if you do not need this facility.

\printAffiliationsAndNotice{}  % leave blank if no need to mention equal contribution
% \printAffiliationsAndNotice{\icmlEqualContribution} % otherwise use the standard text.

% \begin{abstract}
% \cs{The title is factual, but boring. There is not a single new word in the title. Need something more creative.}
% %
% \cs{What is goal of the paper?}
% \cs{Do not use parenthesis, it is for authors who cannot choose what is important.}
% %
% Catastrophic Forgetting is known to be as one of the main challenges existing in continual learning (learning a sequence of tasks). \cs{so?} 
% %
% The stability-plasticity dilemma in an artificial neural network is the main cause of this phenomenon.   Precisely,  current networks tend to be plastic (learning new tasks easily), but they suffer from stability (preserving old tasks’ knowledge) with a bit of change in parameter space.  
% %
% \cs{I would not make the brain connection too strong, not our expertise.}
% In this paper, given the fact that the human brain works in an adaptive manner and data-driven kernel learning provides an adaptive model in machine learning, we rephrase \cs{rephrase does not sound very novel} the continual learning problem as a problem of learning a sequence of kernels. By learning a separate kernel per each task, we hypothesize \cs{vague} that stability property is obtained by minimizing the amount of interference between tasks. 
% %
% \cs{This is the most concrete and clear sentence in the abstract, why keep until the end?}
% This paper provides a technique to incorporate the idea of kernel learning into the continual learning paradigm and studies its impact on catastrophic forgetting.
% \end{abstract}
\begin{abstract}
%Continual Learning is the problem of learning a sequence of tasks that suffers from Catastrophic Forgetting. To overcome this problem, this paper provides a technique to incorporate the idea of data-driven kernel learning into the continual learning paradigm and studies its impact on catastrophic forgetting. The main cause of this problem results from the plasticity-stability dilemma. Current artificial neural network is plastic and they learn new task easily. However, they are inadequate for being stable which means that they forget their previous knowledge. Thanks to adaptability properties of data-driven kernel learning, by learning a separate kernel per each task, we shows that preserving old knowledge is obtained by minimising the amount of interference between tasks. Using this technique, our methods guaranties that both of this objectives, stability and plasticity, are satisfied. We shows that our method is state-of-the-art in Rotated MNIST, Permuted MNIST, and Split CIFAR. 
This paper introduces \textit{kernel continual learning}, a simple but effective variant of continual learning that leverages the non-parametric nature of kernel methods to tackle catastrophic forgetting. We deploy an episodic memory unit that stores a subset of samples for each task to learn  task-specific classifiers based on kernel ridge regression. 
This does not require memory replay and systematically avoids task interference in the classifiers. We further introduce variational random features to learn a data-driven kernel for each task. To do so, we formulate kernel continual learning as a variational inference problem, where a random Fourier basis is incorporated as the latent variable.
The variational posterior distribution over the random Fourier basis is inferred from the coreset of each task. In this way, we are able to generate more informative kernels specific to each task, and, more importantly, the coreset size can be reduced to achieve more compact memory, resulting in more efficient continual learning based on episodic memory. Extensive evaluation on four benchmarks demonstrates the effectiveness and promise of kernels for continual learning. 
\end{abstract}

\section{Introduction}
\label{intro}

%\cs{I prefer to merge the first two paragraphs, so we can talk about kernels in paragraph 2. We can easily drop the AGI part, it is irrelevant, more suited for a blog post.}

%\cs{I would drop this entire first paragraph, too much blabla.}
%Nowadays, with decent progress happening in the field of machine learning, there are a demanding desires to exploit these advancements on edge devices. However, 
% few current state-of-the-art models can be used due to several reasons. 
%current models urge to access quite a bit of training data
%. Moreover, models are 
%and to be trained offline.
%, and this is in contrast with the nature of learning existing in human brain.
%This means that a model having significant performance must be trained over all data simultaneously. 
%These requirements cannot be achieved on edge devices due to the lack of computation power, memory footprint, as well as training and inference time constraints.
%. Furthermore, on edge devices that are working in real-time applications, training model as well as solving the current task at the same time are far-fetched. 
%To mitigate these problems, Continual Learning (CL), also known as Lifelong Leaning or Online Learning, has been introduced~\cite{Lange2019ContinualLA}. 
%This problem definition supports the basic learning mechanism happening in the human brain.

%\cs{The first paragraph does not add much and can also be removed.}
%Humans have the remarkable ability to continually learn new skills, without forgetting past knowledge and experience as acquired over their lifetime. 
Despite the promise of artificially intelligent agents ~\cite{lecun2015deep,schmidhuber2015deep}, they are known to suffer from catastrophic forgetting when learning over non-stationary data distributions ~\cite{McCloskey1989CatastrophicII,Goodfellow2013AnEI}.
Continual learning ~\cite{ring1998child,lopez2017gradient,nguyen2017variational}, also known as life-long learning, was introduced to deal with catastrophic forgetting. In this framework, agent continually learns to solve a sequence of non-stationary tasks by accommodating new information, while remaining able to complete previously experienced tasks with minimal performance degradation. 
%Continual learning has recently attracted increasing attention in machine learning as it is considered one of the desirable proprieties a learning agent should have to achieve artificial general intelligence \cite{hassabis2017neuroscience,hadsell2020embracing}.
%\cs{There is only one citation in this paragraph, from 2016. It seems the topic is dead?}
The fundamental challenge in continual learning is catastrophic forgetting, which is caused by the interference among tasks from heterogeneous data distributions~\cite{Lange2019ContinualLA}. %To overcome catastrophic forgetting, a learning agent needs to strike a balance between stability and plasticity \cite{stability-plasticity}. %That is, it should be plastic to acquire new information while being stable to preserve previously acquired knowledge. 
%How to maximally avoid task interference becomes crucial to alleviate catastrophic forgetting.

% A straightforward way to do this would be  for instance by progressively expanding model parameters \cite{rusu2016progressive}. 
% While being extremely plastic to learn new tasks, this paradigm tends to be infeasible , which results in an unbounded model and prevents useful knowledge transfer across tasks. 
% To enable knowledge sharing, existing works based on neural networks, e.g., use a shared feature extractor and classifier among tasks by imposing constraints on weight update to keep stability or deploying experience replay to conquer forgetting. However, a shared classifier leads to interference in that different tasks have different classes~\cite{nguyen2017variational}. [not complete]
Task interference is almost unavoidable when model parameters, like the feature extractor and the classifier, are shared by all tasks. At the same time, it is practically infeasible to keep a separate set of model parameters for each individual task when learning with an arbitrarily long sequence of tasks \cite{hadsell2020embracing}. Moreover, knowledge tends to be shared and transferred across tasks more in the lower layers than higher layers of deep neural networks \cite{ramasesh2020anatomy}. This has motivated the development of non-parametric classifiers that automatically avoid task interference without sharing any parameters across tasks. Kernel methods \cite{smola1998learning} provide a well-suited tool for this due to their non-parametric nature, and have proven to be a powerful technique in machine learning ~\cite{cristianini2000introduction,smola2004tutorial,rahimi2007random,sinha2016learning}. 
Kernels have been shown to be effective for incremental and multi-task learning with support vector machines \cite{diehl2003svm, pentina2015multi}.
Recently, they have also been demonstrated to be strong learners in tandem with deep neural networks \cite{wilson2016deep,wilson2016stochastic,tossou2019adaptive}, especially when learning from limited data \cite{zhen2020learning,patacchiola2020bayesian}. 
Inspired by the success of kernels in machine learning, we introduce task-specific classifiers based on kernels by decoupling the feature extractor from the classifier for continual learning.

\begin{figure*}[t!]
\centering
\includegraphics[width=1\textwidth]{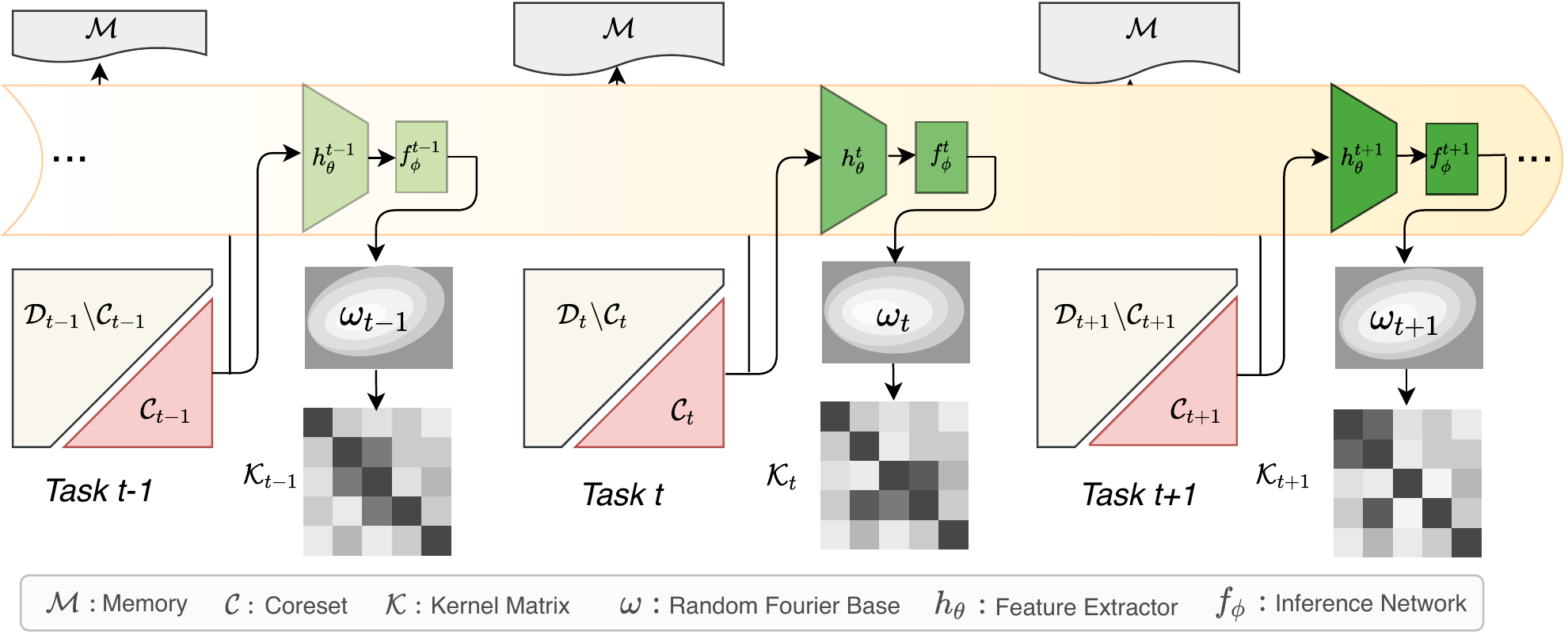}
\caption{\textbf{Overview of kernel continual learning} with variational random features. For each task $t$, we use the coreset $\mathcal{C}_t$ to infer the random Fourier basis, which generates  kernel matrix $\mathcal{K}_t$. The classifier for this task is constructed based on kernel ridge regression using $\mathcal{K}_t$. $h_{\theta}$ denotes the feature extraction network,  parameterized by $\theta$, which is shared and updated when training on the task sequence. $f_{\phi}$ is the inference network, parameterized with $\phi$ for random Fourier bases, which is also shared across tasks and updated throughout learning. Memory $\mathcal{M}$ stores the coreset from each task and is used for inference only. $h_{\theta}$ and $f_{\phi}$ are jointly learned end-to-end.
%\cs{Nice figure. Legend: CoreSet $\xrightarrow{}$ Coreset?}
}
\label{fig:demo}
\end{figure*}

%Generally, CL is the problem of learning a batch of small tasks sequentially. Rephrasing learning problem as a continual learning problem brings about a few challenges. Forgetting previous knowledge (known as Catastrophic Forgetting) during learning a new task can be considered as the most important challenge~\cite{mccloskey1989catastrophic, ratcliff1990connectionist, hebb1949organization}. 
%To explore about why this phenomenon happens, we can judge current neural networks from \textit{Stability-Plasticity} viewpoint. 
%In neural networks, plasticity asserts that models tend to perform better on current data than the previous ones observed. 
%However, they are less stable model which means that they do not provide any guarantees that past knowledge will be untouched. 
% Being unstable about previous knowledge and having tendency to learn new tasks leads to a problem called task interference. 
% To address this problem several methods have been proposed~\cite{OGD, AGEM, shin2017continual, nguyen2017variational}. 
% These methods aim to provide stability solution for current neural network. Meanwhile, they preserve their plasticity property. For instances, in~\cite{AGEM}, a model augmented with memory is introduced to preserve old knowledge by doing replay. This method is used when having memory is allowed. 
% Moreover, \cite{OGD} provides a method to project the gradient of current task over the subspace of previous tasks to acquire stability. 

%\cs{Some parts of paragraph below should go to kernel paragraph above, I cannot follow the replay argument.}
In this paper, we propose \textit{kernel continual learning} to deal with catastrophic forgetting in continual learning. Specifically, we propose to learn non-parametric classifiers based on kernel ridge regression. To do so, we deploy an episodic memory unit to store a subset of samples from the training data for each task, which we refer to as `the coreset', and learn the classifier based on the kernel ridge regression. Kernels provide several benefits. First, the direct interference of classifiers is naturally avoided as kernels are established in a non-parametric way for each task and no classifier parameters are shared across tasks. Moreover, in contrast to existing memory replay methods, e.g.,~\cite{kirkpatrick2017overcoming,AGEM}, our kernel continual learning does not need to replay data from previous tasks when training the current task, which avoids task interference while enabling more efficient optimization. 
In order to achieve adaptive kernels for each task, we further introduce random Fourier features to learn kernels in a data-driven manner. Specifically, we formulate kernel continual learning with random Fourier features as a variational inference problem, where the random Fourier basis is treated as a latent variable. The variational inference formulation naturally induces a regularization term that encourages the model to learn adaptive kernels for each task from the coreset only. As a direct result, we are able to achieve more compact memory, which reduces the storage overhead. 

We perform experiments on four benchmark datasets: Rotated MNIST, Permuted MNIST, Split CIFAR100 and miniImageNet. The results demonstrate the effectiveness and promise of kernel continual learning, which delivers state-of-the-art performance on all benchmarks.

\section{Related Works} \label{related}
%========================================
A fundamental problem in continual learning is catastrophic forgetting. Existing methods differ in the way they deal with this. We will briefly review them in terms of regularization, dynamic architectures and experience replay. For a more extensive overview we refer readers to the reviews by \citet{Parisi2018ContinualLL} and \citet{Lange2019ContinualLA}.

Regularization methods~\cite{kirkpatrick2017overcoming, MAS, lee2017overcoming, zenke2017continual, kolouri2019attention} determine the importance of each model's parameter for each task, which prevents the parameters from being updated for new tasks. \citet{kirkpatrick2017overcoming}, for example, specify the performance of each weight with a Fisher information matrix. Alternatively, \citet{MAS}, determine parameter importance by the gradient magnitude. Naturally, these methods can also be explored from the perspective of Bayesian optimization ~\cite{nguyen2017variational, titsias2019functional, schwarz2018progress, ebrahimi2019uncertainty, ritter2018online}. For instance, \citet{nguyen2017variational} introduce a regularization technique to protect their model against forgetting. Bayesian or not, all these methods address catastrophic forgetting by adding a regularization term to the main loss function. As shown by \citet{Lange2019ContinualLA}, the penalty terms proposed in such algorithms are unable to prevent drifting in the loss landscape of previous tasks. While alleviating forgetting, the penalty term also unavoidably prevents the plasticity to absorb new information from future tasks learned over a long timescale \cite{hadsell2020embracing}. 

Dynamic architectures~\cite{rusu2016progressive, yoon2018lifelong, Jerfel2018ReconcilingMA, li2019learn} allocate a subset of the model parameters for each task. This is achieved by a gating mechanism~\cite{wortsman2020supermasks, Gating}, or by incrementally adding new parameters to the model \cite{rusu2016progressive}. Incremental learning and pruning is another possibility~\cite{PackNet}. Given an over-parameterized model with the ability to learn quite a few tasks, \citet{PackNet} achieve model expansion by pruning the parameters not contributing to the performance of the current task, while keeping them available for future tasks. These methods are preferred when there is no memory usage constraint and the final model performance is prioritized. They offer an effective way to avoid task interference and catastrophic forgetting, but suffer from potentially unbounded model expansion and prevent positive knowledge transfer across tasks.
%\cs{Should we conclude by saying our method benefits from memory expansion, while being able to profit from positive knowledge transfer across tasks? Or just emphasize the knowledge transfer across tasks?}

Experience replay methods~\cite{Lange2019ContinualLA} assume  it is possible to access data from previous tasks by having a fixed-size memory or a generative model able to produce samples from old tasks~\cite{lopez2017gradient, riemer2018learning, rios2018closed, shin2017continual, zhang2019prototype}. \citet{Rebuffi2016iCaRLIC} introduce a model augmented with fixed-size memory, which accumulates samples in the proximity of each class center. \citet{Chaudhry2019OnTE} propose another memory-based model by exploiting a reservoir sampling strategy in the raw input data  selection phase. Rather than storing the original samples, \citet{AGEM} accumulate the parameter gradients during task learning. \citet{shin2017continual} incorporate a generative model into a continual learning model to alleviate catastrophic forgetting by producing samples from previous tasks and retraining the model using data from both previous tasks and the current one.
These papers assume that an extra neural network, such as a generative model, or a memory unit is available. Otherwise, these methods cannot be exploited. 
%
%\cs{This sentence is hard to parse, suggest to split:}
Replay-based methods benefit from a memory unit to retrain their model over previous tasks. In contrast, our proposed method only uses memory to store data as a task identifier proxy at \textit{inference time} without the need of replay for training, which mitigates the optimization cost.
% in memory based methods. 

%%methodology
% \begin{figure*}[ht!]
% \centering
% \begin{subfigure}{.5\textwidth}
%   \centering
%   \includegraphics[width=.99\linewidth]{figs/KCL.pdf}
%   \caption{Network 1 (Stable Network)}
%   \label{fig:appendix-intro-stable}
% \end{subfigure}\hspace{0.05\textwidth}
% \begin{subfigure}{.4\textwidth}
%   \centering
%   \includegraphics[width=.99\linewidth]{figs/inference_model.pdf}
%   \caption{Network 2 (Plastic Network)}
%   \label{fig:appendix-intro-plastic}
% \end{subfigure}
% \caption{Full results for the comparison of accuracy on all tasks for two networks in Figure~\ref{fig:intro} of the introduction section.}
% \label{fig:appendix-intro-all-tasks}
% \end{figure*}

%========================================
\section{Kernel Continual Learning}
%========================================
%In the first section, problem statement, we rephrase continual learning problem as  a kernel continual learning problem. In the next parts, Random Fourier Features is restated, and the loss function for our work will be introduced.  
\subsection{Problem Statement}

% \cs{Shouldn't this first paragraph be presented in the introduction?}
% % Define classification problem
% In a traditional classification problem, a model or agent is learned to map input data to its corresponding output space like $M:X \mapsto Y$. It is assumed that all input-output pairs are available during model training. However, in practice, the situation is often different. Input-output pairs are observed as a stream of data, and a model is learned incrementally. This training regime is called continual learning. More precisely, we observe training data as a sequence of different tasks, $T_1, T_2, \cdot \cdot \cdot T_n$ where $n$ equals the number of tasks, and each task is a separate classification problem. 
% %
% \cs{I don't understand:}
% Whenever each task is observed, it is no longer available in the future, and we are unable to re-train the model over previous tasks. 
In the traditional supervised learning setting, a model or agent $f$ is learned to map input data from the input space to its target in the corresponding output space: $\mathcal{X} \mapsto \mathcal{Y}$, where samples $X \in \mathcal{X}$ are assumed to be drawn from the same data distribution. In the case of the image classification problem, $X$ are the images and $Y$ are associated class labels. Instead of solving a single task, continual learning aims to solve a sequence of different tasks, $T_1, T_2, \cdot \cdot \cdot T_n$, from non-stationary data distributions, where $n$ stands for the number of tasks, and each of which is an individual classification problem. A continual learner is required to continually solve each $t$ of those tasks once being trained on its labeled data, while remaining able to solve previous tasks with no or limited access to their data.

Generally, a continual learning model based on a neural network is comprised of a feature extractor $h_\theta$ and a classifier $f_c$. The feature extractor is a convolutional architecture found before the last fully connected layer, which is shared across tasks. The classifier is the last fully connected layer. We propose to learn a task-specific, non-parametric classifier based on kernel ridge regression.

We consider learning the model on the current task $t$. Given its training data $\mathcal{D}_t$, we choose uniformly a subset of data between existing classes in current task $t$, which is called the \textit{coreset dataset} \cite{nguyen2017variational} and denoted as: $\mathcal{C}_t{=}{(\x_i, \y_i)}_{i{=}1}^{N_c}$. We construct the classifier $f_c$ based on kernel ridge regression on the coreset. Assume we have the classifier with weight $\mathbf{w}$, and the loss function of kernel ridge regression takes the following form:
\begin{equation}
    \mathcal{L}_{\rm{krr}}(\mathbf{w}) = \frac{1}{2}\sum_i(\y_i - \mathbf{w}^\top \psi(\x_i)) + \frac{1}{2}\lambda ||\mathbf{w}||^2,
\end{equation}
where $\lambda$ is the weight decay parameter.
Based on the Representer theorem \cite{scholkopf2001generalized}, we have: 
\begin{equation}
   \mathbf{w} = f^{\alpha^t}_c(\cdot) = \sum^{N_c}_{i=1} \alpha_i k(\cdot,\psi(\x_i)),
\end{equation}
where $k(\cdot,\cdot)$ is the kernel function. Then $\alpha$ can be calculated in a closed form:
\begin{equation}
    \alpha^t = Y(\lambda I + \mathcal{K})^{-1},
\end{equation}
where $\alpha^t {=} [\alpha_1, \cdots, \alpha_i, \cdots, \alpha_{N_C}]$ and $\lambda$ is considered to be a learnable hyperparameter. The $\mathcal{K} \in R^{N_c \times N_c}$ matrix for each task is computed as $k(\x_i, \x_j) {=} \psi(\x_i)\psi(\x_j)^\top$. Here, $\psi(\x_i)$ is the feature map of $\x_i \in \mathcal{C}_t$, which can be obtained from the feature extractor $h_\theta$.
%, and \mo{$\x_i$ and $\x_j$ are from $\mathcal{C}_t$.}

%and $R_t{=}{(x_i, y_i)}_{i{=}1}^{N-N_c}$ for a task, 

% the parameters ($\alpha^t$) of the predictor function ($F_{a^t}$) are estimated using a standard learning algorithms by the kernel trick $\alpha^t{=}\Lambda(\Phi(X), Y)$. 
% Here, $\Lambda$ is the base-learner; $\Phi: X \mapsto R^X$ provide a mapping function from input space (X) to a Hilbert (H) space, or dot product space (For more information about kernel learning, see~\cite{hofmann2008kernel}); and $Y$ is the target space.

%To determine the base-learner for task $t$, kernel functions such as radius basis function and polynomial base function can be utilized. These base functions map the input data point into a Hilbert space in an efficient manner. After obtaining the base-learner on task $t$ using $C_t$, a base-learner's performance is evaluated on $R_t$ by the following loss function:

To jointly learn the feature extractor $h_\theta$, we minimize the overal loss function over samples from the remaining set:
\begin{equation}
    \sum_{(\x', \y') \in D_t\backslash \mathcal{C}_t}
    {\mathcal{L}(f^{\alpha^t}_{c}({\psi(\x')), \y'})}.
\end{equation}
Here, we choose $\mathcal{L}(\cdot)$ to be the cross-entropy loss function and the predicted output $\tilde{\y}'$ is computed by
% \begin{equation}
%     \sum_{t}^{n} \sum_{(\tilde{\mathbf{x}}, \tilde{\mathbf{y}}) \in \mathcal{R}^{t}} \mathcal{L}\left(f_{\alpha^{t}}\left(\Phi^{t}(\tilde{\mathbf{x}})\right), \tilde{\mathbf{y}}\right), \text { s.t. } \alpha^{t}=\Lambda\left(\Phi^{t}(X), Y\right)
% \end{equation}
% For each task, one mapping function ($\Phi^t$) must be achieved using the above loss function. To do so, we need to specify a specific kernel function per task ($k^t$). Among different methods existing in the kernel trick literature (see~\cite{hofmann2008kernel}), random Fourier features would be an appropriate choice because of its efficiency.
% For the classifier part of our model, $\Lambda$, we rely on a kernel-ridge regression model that has a closed-form solution.
%  To obtain the base-learner $\Lambda$ for task $t$, the following optimization problem should be solved w.r.t. $C_t$:
% \begin{equation}
%     \Lambda=\underset{\alpha}{\arg \min } \operatorname{Tr}\left[(Y-\alpha K)(Y-\alpha K)^{\top}\right]+\lambda \operatorname{Tr}\left[\alpha K \alpha^{\top}\right]
% \label{loss:ridge-opt}
% \end{equation}
% After finding the base-learner using the coreset dataset ($C_t$) for a given task, 
% the model prediction on the remaining dataset ($R_t$) is computed as:
\begin{equation}
  \tilde{\y}' = f^{\alpha^t}_c(\psi(\x')) = \texttt{Softmax}(Y(\lambda I + \mathcal{K})^{-1} \tilde{K}),
  \label{eq:prediction}
\end{equation}
where $\tilde{K} {=} \psi(X)\psi(\x')^{\top}$, $\psi(X)$ denotes the feature maps of samples in the coreset, and $\texttt{Softmax}(\cdot)$ is the softmax function applied to the output of the kernel ridge regression.

In principle, we can use any (semi-)positive definite kernel, e.g., a radial basis function (RBF) kernel or a dot product linear kernel to construct the classifier. However, none of those kernels are task specific, potentially suffering from suboptimal performance, especially with limited data. Moreover, we would require a relatively large coreset to obtain informative and discriminative kernels for satisfactory performance. To address this, we further introduce random Fourier features to learn data-driven kernels, which have previously demonstrated success in regular learning tasks \cite{bach2004multiple,sinha2016learning,carratino2018learning,zhen2020learning}. Data-driven kernels using random Fourier features provides an appealing technique to learn strong classifiers with a relatively small memory footprint for continual learning based on episodic memory.

\subsection{Variational Random Features}
One of the key ingredients when finding a mapping function in non-parametric approaches, such as kernel ridge regression, is the kernel function. \citet{rahimi2007random} introduced an algorithm to approximate translation-invariant kernels using explicit feature maps, which is theoretically underpinned by Bochner's theorem~\cite{rudin1962fourier}.

\begin{theorem}[Bochner's Theorem]
\label{theorem1}
\textit{A continuous, real valued, symmetric and shift-invariant function $\mathrm{k}\left(\mathbf{x}, \mathbf{x}^{\prime}\right)=\mathrm{k}(\mathbf{x}-\mathbf{x}^{\prime})$ on $\mathbb{R}^\mathrm{d}$ is a positive definite kernel if and only if it is the Fourier transform $p(w)$ of a positive finite measure such that:}
\begin{equation}
\begin{aligned}
\mathrm{k}\left(\mathbf{x}, \mathbf{x}^{\prime}\right)=& \int_{\mathbb{R}^{d}} e^{i \omega^{\top}\left(\mathbf{x}-\mathbf{x}^{\prime}\right)} d p(\boldsymbol{\omega})=\mathbb{E}_{\omega}\left[\zeta_{\omega}(\mathbf{x}) \zeta_{\omega}\left(\mathbf{x}^{\prime}\right)^{*}\right] \\
& \text { where } \zeta_{w}(\mathbf{x})=e^{i \omega^{\top} \mathbf{x}}.
\end{aligned}
\end{equation}
\end{theorem}
With a sufficient number of samples $\omega$ drawn from $p(\omega)$, we can achieve an unbiased estimation of $\mathrm{k}\left(\mathbf{x}, \mathbf{x}^{\prime}\right)$  by  $\zeta_{w}(\mathbf{x})\zeta_{w}(\mathbf{x})^*$ \cite{rahimi2007random}.

Based on Theorem~\ref{theorem1}, we draw $D$ sets of samples: $\{{\omega_i}\}_{i=1}^{D}$ and $\{{b_i}\}_{i=1}^{D}$ from a normal distribution and uniform distribution (with a range of [$0$, $2\pi$]), respectively, and construct the random Fourier features (RFFs) for each data point $\x$ using the formula: 
\begin{equation}
\psi(\mathbf{x})=\frac{1}{\sqrt{D}}\left[\cos \left(\boldsymbol{\omega}_{1}^{\top} \mathbf{x}+b_{1}\right), \cdots, \cos \left(\boldsymbol{\omega}_{D}^{\top} \mathbf{x}+b_{D}\right)\right].
\label{eq:rff}
\end{equation}
Having the random Fourier features,  we calculate the kernel matrix by $\mathrm{k}\left(\mathbf{x}, \mathbf{x}^{\prime}\right) {=} \psi(\x)\psi(\x^\prime)^\top$.

Traditionally the shift-invariant kernel is constructed based on random Fourier features, where the Fourier basis is drawn from a Gaussian distribution transformed from a pre-defined kernel. This results in kernels that are agnostic to the task. In continual learning, however, tasks are provided sequentially from non-stationary data distributions, which makes it suboptimal to share the same kernel function across tasks. To address this problem, we propose to learn task-specific kernels in a data-driven manner. This is even more appealing in continual learning as we would like to learn informative kernels using a coreset of a minimum size. We formulate it as a variational inference problem, where we treat the random basis $\omega$ as a latent variable. 

%Precisely, RFFs are constructed using samples resulting from a variational posterior distribution. 

%\xz{The coreset $\mathcal{C}$ is supposed to be in the variational posterior, because the E term is taken over the variational posterior. This is also what you implemented: using the coreset to compute the kernel for classification.}

\paragraph{Evidence Lower Bound} From the probabilistic perspective, we would like to maximize the following conditional predictive log-likelihood for the current task $t$:
\begin{equation}
\max _{p} \sum_{(\mathbf{x}, \mathbf{y}) \in \mathcal{D}_t\backslash \mathcal{C}_t} \ln p(\mathbf{y} | \mathbf{x}, \mathcal{D}_t\backslash \mathcal{C}_t),
\label{cll}
\end{equation}
which amounts to making maximally accurate predictions on $\x$ based on $\mathcal{D}_t\backslash \mathcal{C}_t$.

By introducing the random Fourier basis $\omega$ into Eq.~(\ref{cll}), which is treated as a latent variable, we have:
\begin{equation}
\max _{p} \sum_{(\mathbf{x}, \mathbf{y}) \in \mathcal{D}_t\backslash \mathcal{C}_t} \ln \int p(\mathbf{y} |\mathbf{x}, \omega, \mathcal{D}_t\backslash \mathcal{C}_t) p_{\gamma}(\omega |\mathcal{D}_t\backslash \mathcal{C}_t)d\omega.
\label{formula:loglikelihood}
\end{equation}
The intuition is that we can use data to infer the distribution over the latent variable $\omega$ whose prior is conditioned on the data. We combine the data and $\omega$ to generate kernels to classify $\x$ based on kernel ridge regression. We can also simply place an uninformative prior of a standard Gaussian distribution over the latent variable $\omega$, which will be investigated in our experiments.

%We naturally obtain a conditional prior that depends on the data, which is in contrast to the unsupervised learning models, e.g., variational auto-encoder using an uninformative isotriphic Gaussian distribution \cite{kingma2013auto, rezende2014stochastic}.

%we incorporate a prior distribution conditioned over batch of samples in $R_t$ and all samples in $C_t$. This prior is adopted from CVAE (conditional variational auto-encoder) (Sohen et. al., 2015). 

It is intractable to directly solve for the true posterior $p(\omega|\x,\y,\mathcal{D}_t\backslash \mathcal{C}_t)$ over $\omega$. We therefore introduce a variational posterior $q_{\phi}(\omega|\mathcal{C}_t)$ conditioned solely on the coreset $\mathcal{C}_t$ because the coreset will be stored as episodic memory for the inference of each corresponding task.

By incorporating the variational posterior into Eq.~(\ref{formula:loglikelihood}) and applying Jensen's inequality, we establish the evidence lower bound (ELBO) as follows:
\begin{equation}
\begin{aligned}
\ln p(\mathbf{y} | \mathbf{x}, \mathcal{D}_t\backslash \mathcal{C}_t)  & \geq \mathbb{E}_{q_{\phi}(\boldsymbol{\omega} | \mathcal{C}_t)}  \big[\ln p(\mathbf{y} |\mathbf{x}, \boldsymbol{\omega},\mathcal{D}_t\backslash \mathcal{C}_t) \big]\\
&-D_{\mathrm{KL}}\big[q_{\phi}(\boldsymbol{\omega} | \mathcal{C}_t) \| p_{\gamma}(\boldsymbol{\omega}|\mathcal{D}_t\backslash \mathcal{C}_t)\big]\\
&=\mathcal{L}_{\mathrm{ELBO}}.
\label{elbo}
\end{aligned}
\end{equation}
Therefore, maximizing the ELBO amounts to maximizing the conditional log-likelihood in Eq.~(\ref{cll}). The detailed derivation is provided in the supplementary materials.

%\xz{The training set (excluding the coreset) should be in the prior to be consistent with your implementation.}

\paragraph{Empirical Objective Function}
In the continual learning setting, we would like the model to be able to make predictions based solely on the coreset $\mathcal{C}_t$ stored in the memory. That is, the conditional log-likelihood should be conditioned on the coreset only. Based on the ELBO in Eq.~(\ref{elbo}) we establish the following empirical objective function which, is minimized by our overall training procedure:
\begin{equation}
\begin{aligned}
\tilde{\mathcal{L}}_{\mathrm{ELBO}} & = \frac{1}{T} \sum _{t=1} ^ {T}\Big[\sum_{(\mathbf{x}, \mathbf{y}) \in \mathcal{D}_t\backslash \mathcal{C}_t} \frac{1}{L}\sum^L_{\ell=1}  \big[\ln p(\mathbf{y} | \mathbf{x}, \boldsymbol{\omega}^{(\ell)},\mathcal{C}_t)\big] \\
&-D_{\mathrm{KL}}\big[q_{\phi}(\boldsymbol{\omega} | \mathcal{C}_t) \| p_{\gamma}(\boldsymbol{\omega} | \mathcal{D}_t\backslash \mathcal{C}_t)\big]\Big],
\end{aligned}
\label{objective}
\end{equation}
where, in the first term, we employ the Monte Carlo method to draw samples from the variational posterior $q(\omega|\mathcal{C}_t)$ to estimate the log-likelihood, and $L$ is the number of Monte Carlo samples. In the second term, the conditional prior serves as a regularizer that ensures the inferred random Fourier basis will always be relevant to the current task. Minimizing the Kullback Leibler (KL) divergence forces the distribution of random Fourier bases, as inferred from the coreset, to be close to the one from the training set. Moreover, the KL term enables us to generate informative kernels adapted to each task using relatively small memory. 

In practice, the conditional distributions $q_{\phi}(\boldsymbol{\omega} | \mathcal{C}_t) $ and $p_{\gamma}(\boldsymbol{\omega} | \mathcal{D}_t\backslash \mathcal{C}_t)$ are assumed to be Gaussian. We implement them by using the amortization technique \cite{kingma2013auto}. That is, we use multilayer perceptrons to generate the distribution parameters, $\mu$ and $\sigma$, by taking the conditions as input. In our experiments, we deploy two separate amortization networks, referred to as the inference network $f_{\phi}$ for the variational posterior and the prior network $f_\gamma$ for the prior. %, for $q_{\phi}(\boldsymbol{\omega} | \mathcal{C}_t) $ and $p_{\gamma}(\boldsymbol{\omega} | \mathcal{D}_t\backslash \mathcal{C}_t)$.
In addition, to demonstrate the effectiveness of data-driven kernels, we also implement a variant of variational random features by replacing the conditional prior in Eq. (\ref{objective}) with an uninformative one, i.e., an isotropic Gaussian distribution $\mathcal{N}(0,\rm{I})$. In this case, kernels are also learned in a data-driven way from the coreset without being regulated by the training data from the task.

\begin{figure*}[t!]
% \vspace{-3mm}
\centering
\begin{subfigure}{.33\textwidth}
  \centering
  \includegraphics[width=.99\linewidth]{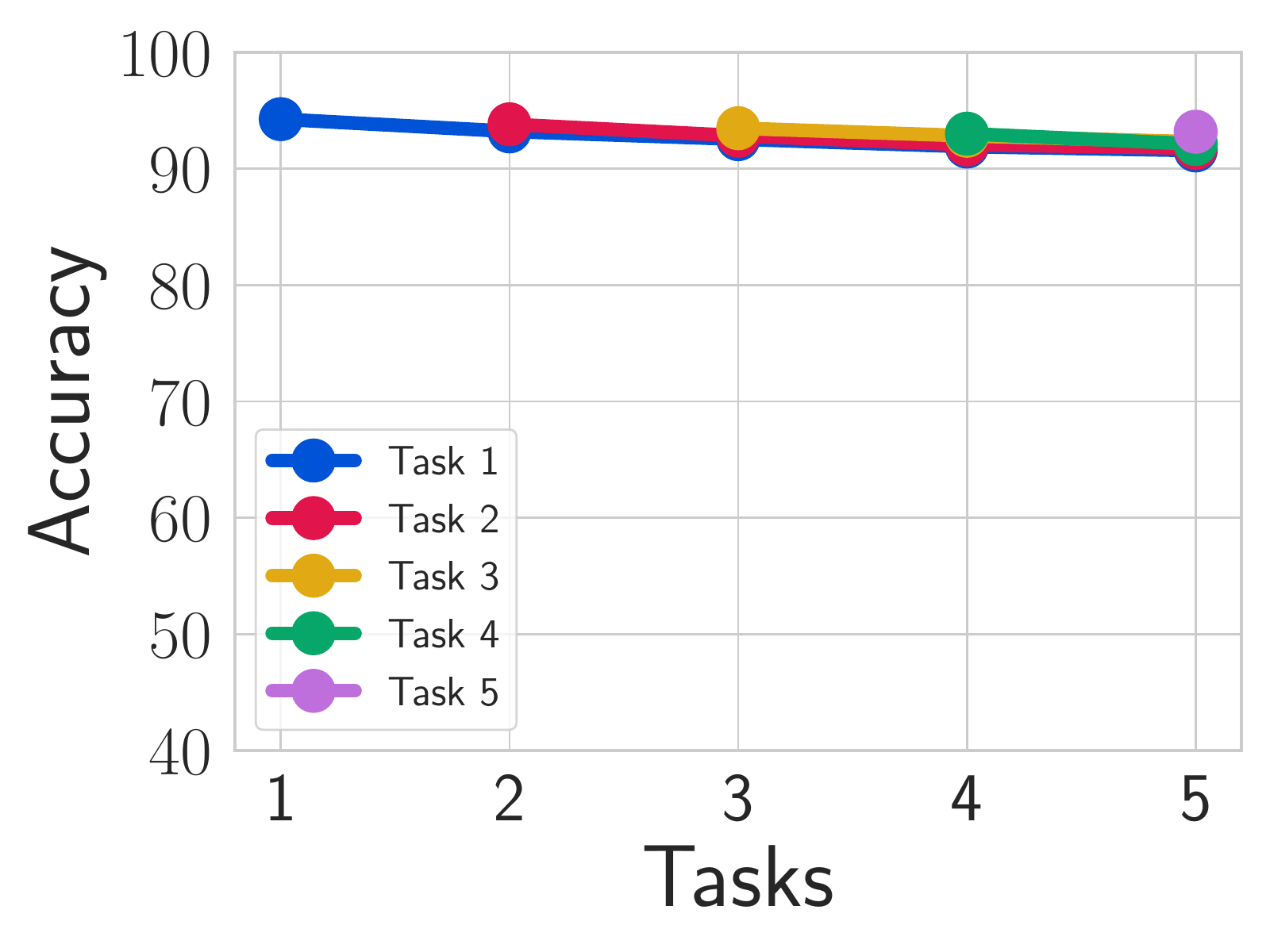}
  \caption{Permuted MNIST}
  \label{fig:5-run-5-task-acc-pm}
\end{subfigure}%
\begin{subfigure}{.33\textwidth}
  \centering
  \includegraphics[width=.99\linewidth]{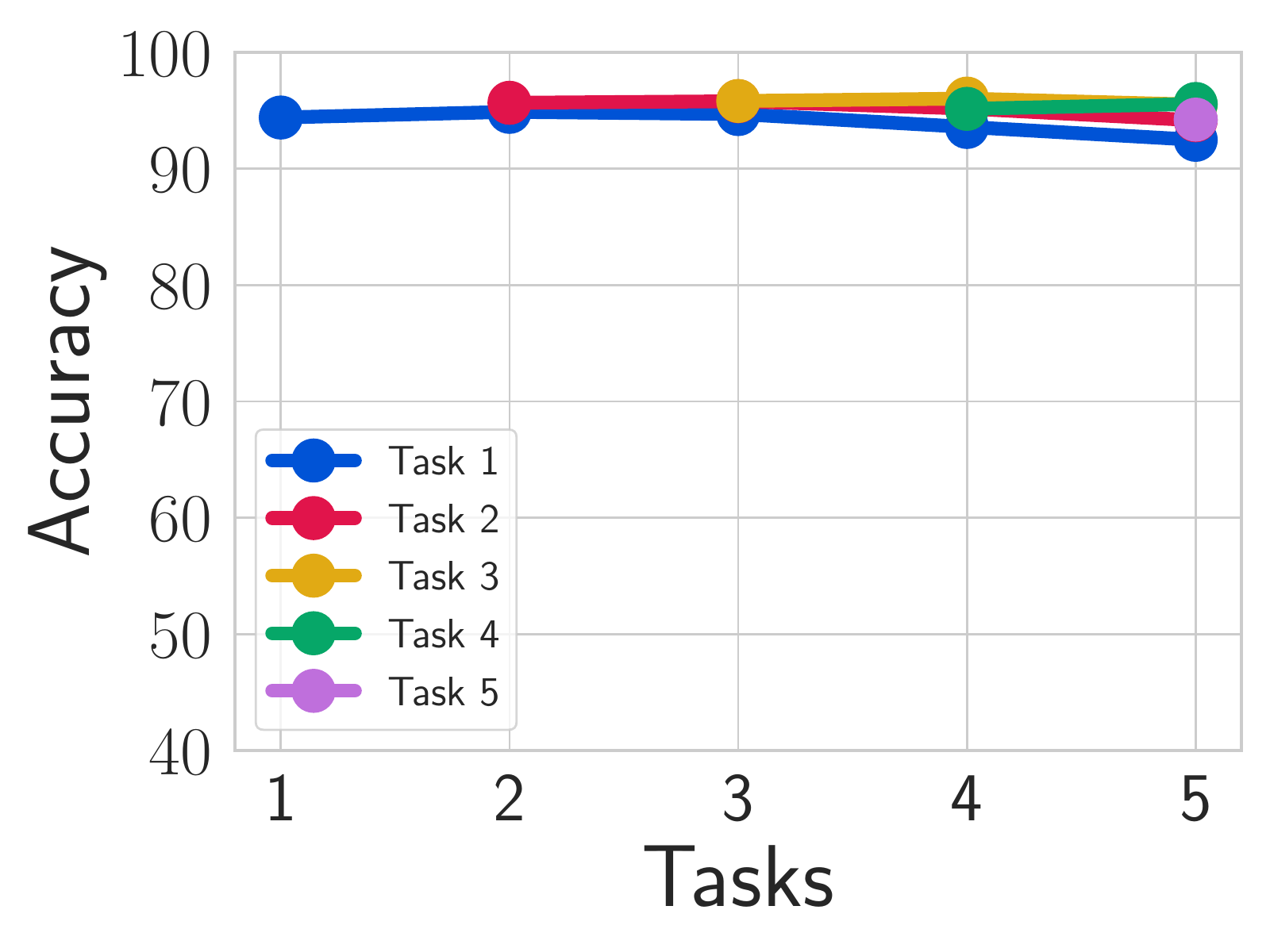}
  \caption{Rotated MNIST}
  \label{fig:5-run-5-task-acc-rm}
\end{subfigure}
\begin{subfigure}{.33\textwidth}
  \centering
  \includegraphics[width=.99\linewidth]{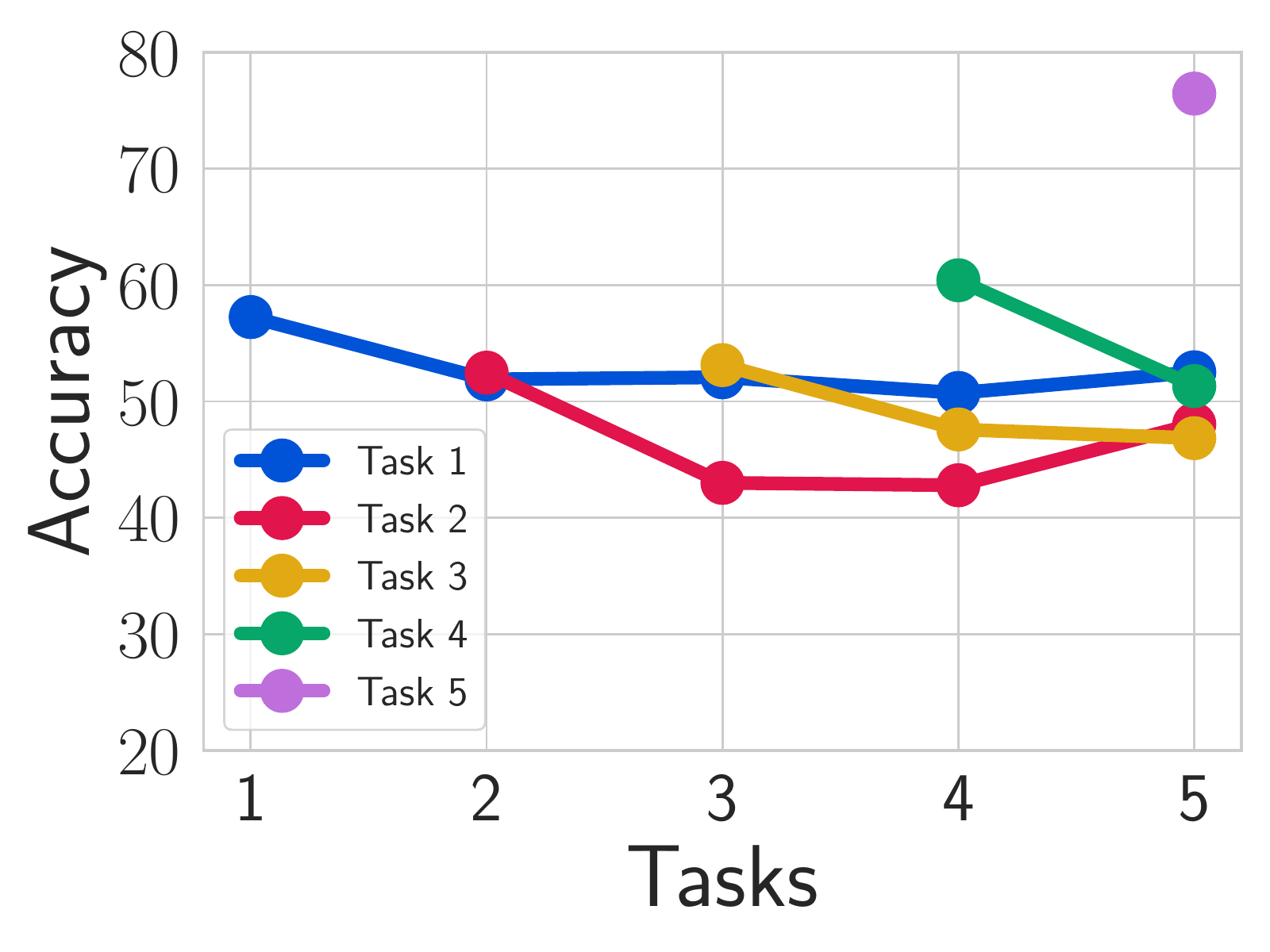}
  \caption{Split CIFAR100}
  \label{fig:5-run-5-task-acc-cifar}
\end{subfigure}
% \vspace{-2mm}
\caption{\textbf{Effectiveness of kernels.} Accuracy of kernel continual learning by variational random features for the first five tasks on three benchmarks. Note the limited decline in accuracy as the number of tasks increase.}
\label{fig:5-run-5-task-acc}
\vspace{-4mm}
\end{figure*}

%========================================
\section{Experiments}
%========================================
We conduct our experiments on four benchmark datasets for continual learning. We perform thorough ablation studies to demonstrate the effectiveness of kernels for continual learning as well as the benefit of variational random features in learning data-driven kernels.
%The experimental results emphasize the effectiveness of the kernel continual learning. Our proposed data-driven kernel learning via variational inference outperforms other predefined kernels.
\subsection{Datasets}

\textbf{Permuted MNIST} Following~\cite{kirkpatrick2017overcoming}, we generate 20 different MNIST datasets. Each dataset is created by a special pixel permutation of the input images, without changing their corresponding labels. Each dataset has its own permutation by owning a random seed.

\textbf{Rotated MNIST} Similar to Permuted MNIST, Rotated MNIST has 20 tasks \cite{mirzadeh2020understanding}. Each task's dataset is a specific random rotation of the original MNIST dataset, 
%
% \cs{Difficulty parsing this sentence:}
e.g., the dataset for task $1$, task $2$, and task $3$ are the original MNIST dataset, a 10-degree rotation, and a 20-degree rotation, respectively. 
In other words, each task's dataset is a 10-degree rotation of the previous task's dataset.

\textbf{Split CIFAR100}  \citet{zenke2017continual} created this benchmark by dividing the CIFAR100 dataset into 20 sections. Each section represents $5$ out of $100$ labels (without replacement) from CIFAR100. Hence, it contains $20$ tasks and each task is a $5$-way classification problem. 

\textbf{Split miniImageNet}  Similar to Split CIFAR100, the miniImageNet benchmark~\cite{vinyals2016matching} contains 100 classes, and is a subset of the original ImageNet dataset~\cite{russakovsky2015imagenet}. It has 20 disjoint tasks, each of which task contains 5 classes.

\subsection{Evaluation Metrics}
We follow the common conventions in continual learning ~\cite{Chaudhry2018RiemannianWF, mirzadeh2020understanding}, and report the \textit{average accuracy} and \textit{average forgetting} metrics. 

\textbf{Average Accuracy} This score shows the model accuracy after training over $t$ consecutive tasks are finished. It is formulated as follows:
\begin{equation}
    \mathrm{A}_t = \frac{1}{t} \sum_{i=1}^{t} \mathrm{a}_{t,i},
    \label{eq:avg_per}
\end{equation}
where $\mathrm{a}_{t,i}$ refers to the model performance on task $i$ after being trained on task $t$.

\textbf{Average Forgetting}
This metric measures the decline in accuracy for each task, according to the highest accuracy and the final accuracy reached after model training is finished. It is formulated as follows:
\begin{equation}
    \mathrm{F} = \frac{1}{T-1} \sum _{i=1} ^{T-1} \rm{max}_{1,...,T-1}(\mathrm{a}_{t,i} - \mathrm{a}_{T,i}).
    \label{eq:avg_forg}
\end{equation}
Taken together, the two metrics allow us to asses how well a continual learner achieves its classification target while overcoming forgetting.

\subsection{Implementation Details}
Our kernel continual learning contains three networks: a shared backbone $h_\theta$, a posterior network $f_\phi$, and a prior network $f_\gamma$. An overview of our implementation is provided in the supplementary materials. For the Permuted MNIST and Rotated MNIST benchmarks, $h_\theta$ contains only two hidden layers, each of which has $256$ neurons, followed by a ReLU activation function. For Split CIFAR100, we use a ResNet18 architecture similar to~\citet{mirzadeh2020understanding}, and for miniImageNet, we have a ResNet18 similar to ~\citet{chaudhry2020continual}. With regard to the $f_\gamma$ and $f_\phi$ networks, we adopt three hidden layers followed by an ELU activation function~\cite{gordon2018meta}. The number of neurons in each layer depends on the benchmark. On Permuted MNIST and Rotated MNIST, there are $256$ neurons per layer, and we use $160$ and $512$ for Split CIFAR100 and miniImageNet, respectively.
For fair comparisons, the model is trained for only \textit{one} epoch per task, that is, each sample in the dataset is observed only once. The batch size is set to 10. Other optimization techniques, such as weight-decay, learning rate decay, and dropout are set to the same values as in \cite{mirzadeh2020understanding}. The model is implemented in Pytorch \cite{Pytorch}. 
% \cs{Include github link here?}
All our code will be released.\footnote{
\href{https://github.com/mmderakhshani/KCL}{https://github.com/mmderakhshani/KCL}
}

%========================================
\subsection{Results}
%========================================
We first provide a set of ablation studies for our proposed method. Then, the performance of our method is compared against other continual learning methods (see the supplementary materials for more details about each ablation).

\paragraph{Effectiveness of kernels} 
To demonstrate the effectiveness of kernels for continual learning, we establish classifiers based on kernel ridge regression using commonly used linear, polynomial, radial basis function (RBF) kernels, and our proposed variational random Fourier features. We report results on Split CIFAR100, where we sample five different random seeds. For each random seed, the model is trained over different kernels. Finally, the result for each kernel is estimated by averaging over the corresponding random seeds. For fair comparison, all kernels are computed using the same coreset of size $20$.

The results are shown in Table \ref{tab:kernel-ablation}. All kernels perform well: the radial basis function (RBF) obtains a modest average accuracy in comparison to other basic kernels such as the linear and polynomial kernels. The linear and polynomial kernels perform similarly. The kernels obtained from variational random features (VRFs) achieve the best performance in comparison to other kernels, and they work better than its uninformative counterpart. This emphasizes that the prior incorporated in VRFs is more informative because its prior is data-driven. 

Regarding VRFs, Figure~\ref{fig:5-run-5-task-acc} demonstrates the change of each task's accuracy on Permuted MNIST, Rotated MNIST and Split CIFAR100. 
% \cs{So? Why is this of interest for the reader? What's the point?}
% \cs{I do not understand this sentence, and how it can be derived from the Table}}
% Increasing batch size improves VRF by providing more informative prior. 
%However, for being consistent with baselines, we use the same batch size 10 as \cite{mirzadeh2020understanding}. 
It is also worth mentioning that the classifiers based on those kernels are non-parametric, enabling them to systematically avoid task interference in classifiers. Thanks to the non-parametric nature of the classifiers based on kernels, our method is flexible and able to naturally deal with a more challenging setting under a different numbers of classes (which we refer to as `varied ways'). To demonstrate this, we conduct experiments with a varying number of classes in each task using VRFs. The results on Split CIFAR100 and Split miniIMageNet 
are shown in Table~\ref{tab:variable_lenght}. Kernel continual learning results in slightly lower accuracy on Split CIFAR100, but leads to an improvement over the traditional fixed ways evaluation on Split miniImageNet. %\cs{So what does this experiment shows? Without comparison to alternative method that collapses in varied way setting, we do not make a point.} \xz{We can remove this table if we have no space}

%\cs{Need to compare to alternative method that is more brittle under this `varied ways' scenario, else the point we want to make is unclear.}

\begin{table}[t]
\centering
\caption{\textbf{Effectiveness of kernels} on Split CIFAR100. 
All kernels perform well, but the simple linear kernel performs better than the RBF kernel. The adaptive kernels based on the random Fourier features achieve the best performance, indicating the advantage of data-driven kernels. 
} %\cs{Note the two VRFs are comparable, difference is within variance.}\cs{Forgetting is close to 0, why do we need to report it? Not meaningful} We can see this after showing this.
\vspace{2mm}
\resizebox{1\columnwidth}{!}{%
\begin{tabular}{@{}lcc@{}}
\toprule
 & \multicolumn{2}{c}{\textbf{Split CIFAR100}} \\ \cmidrule(lr){2-3} 
\textbf{Kernel} & Accuracy & Forgetting  \\ \midrule
RBF         & 56.86 {\scriptsize $\pm$ 1.67} & 0.03 {\scriptsize $\pm$ 0.008} \\
Linear      & 60.88 {\scriptsize $\pm$ 0.64} & 0.05 {\scriptsize $\pm$ 0.007} \\
Polynomial  & 60.96 {\scriptsize $\pm$ 1.19} & 0.03 {\scriptsize $\pm$ 0.004} \\
VRF (uninformative prior) &  62.46 {\scriptsize $\pm$ 0.93} & 0.05 {\scriptsize $\pm$ 0.004} \\
VRF         & 62.70 {\scriptsize $\pm$ 0.89} & 0.06 {\scriptsize $\pm$ 0.008} \\
\bottomrule
\end{tabular}%
}
\label{tab:kernel-ablation}
\end{table}

\begin{table}[t!]
\centering
\caption{\textbf{Effectiveness of VRF kernel} for variable-way scenario on Split CIFAR100 and Split miniImageNet. In this scenario, instead of covering a fixed number of five classes per task from Split CIFAR100 and Split miniImageNet, a task is able to cover a more flexible number of classes in the range $[3, 15]$. By doing so, the experimental setting is more realistic. Even in this case, our proposed method is effective, as indicated by the performance improvement of miniImageNet.}
\vspace{2mm}
\resizebox{1\columnwidth}{!}{%
\begin{tabular}{lllll}
\toprule
 & \multicolumn{2}{c}{\textbf{Split CIFAR100}} & \multicolumn{2}{c}{\textbf{Split miniImageNet}} \\ \cmidrule(lr){2-3} \cmidrule(lr){4-5}
 & Accuracy & Forgetting & Accuracy & Forgetting  \\ \midrule 
Fixed Ways  & 64.02 & 0.05  & 51.89 & 0.06 \\
Varied Ways & 61.00\scriptsize{$\pm$ 1.80} & 0.05\scriptsize{$\pm$ 0.01} & 53.90\scriptsize{$\pm$ 2.95} & 0.05\scriptsize{$\pm$ 0.01} \\
\bottomrule
\end{tabular}%
}
\label{tab:variable_lenght}
\end{table}

\begin{figure*}[t!]
% \vspace{-3mm}
\centering
\begin{subfigure}{.33\textwidth}
  \centering
  \includegraphics[width=.99\linewidth]{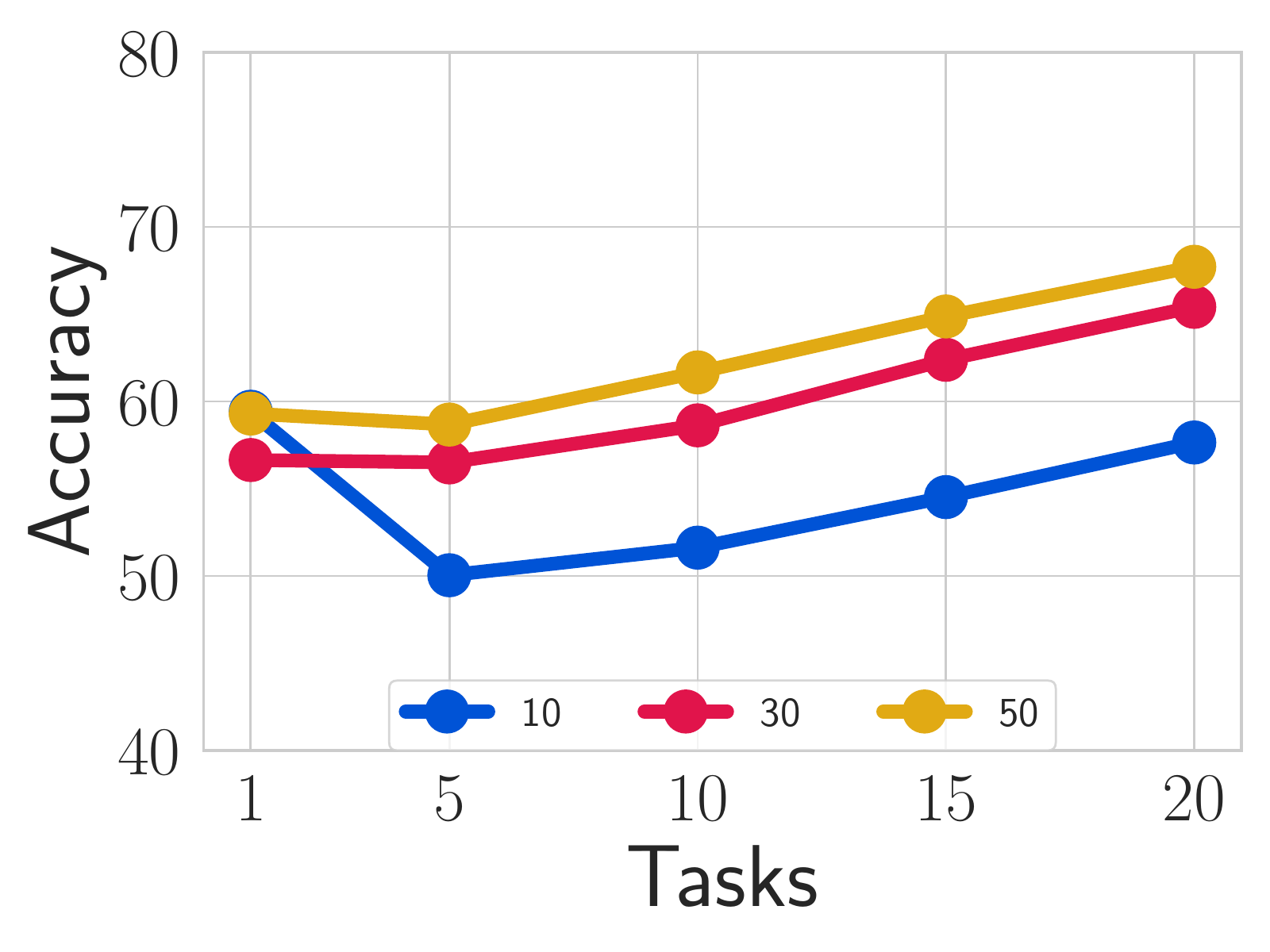}
  \caption{Split CIFAR100}
  \label{fig:positive_transfer_cifar}
\end{subfigure}%
\begin{subfigure}{.33\textwidth}
  \centering
  \includegraphics[width=.99\linewidth]{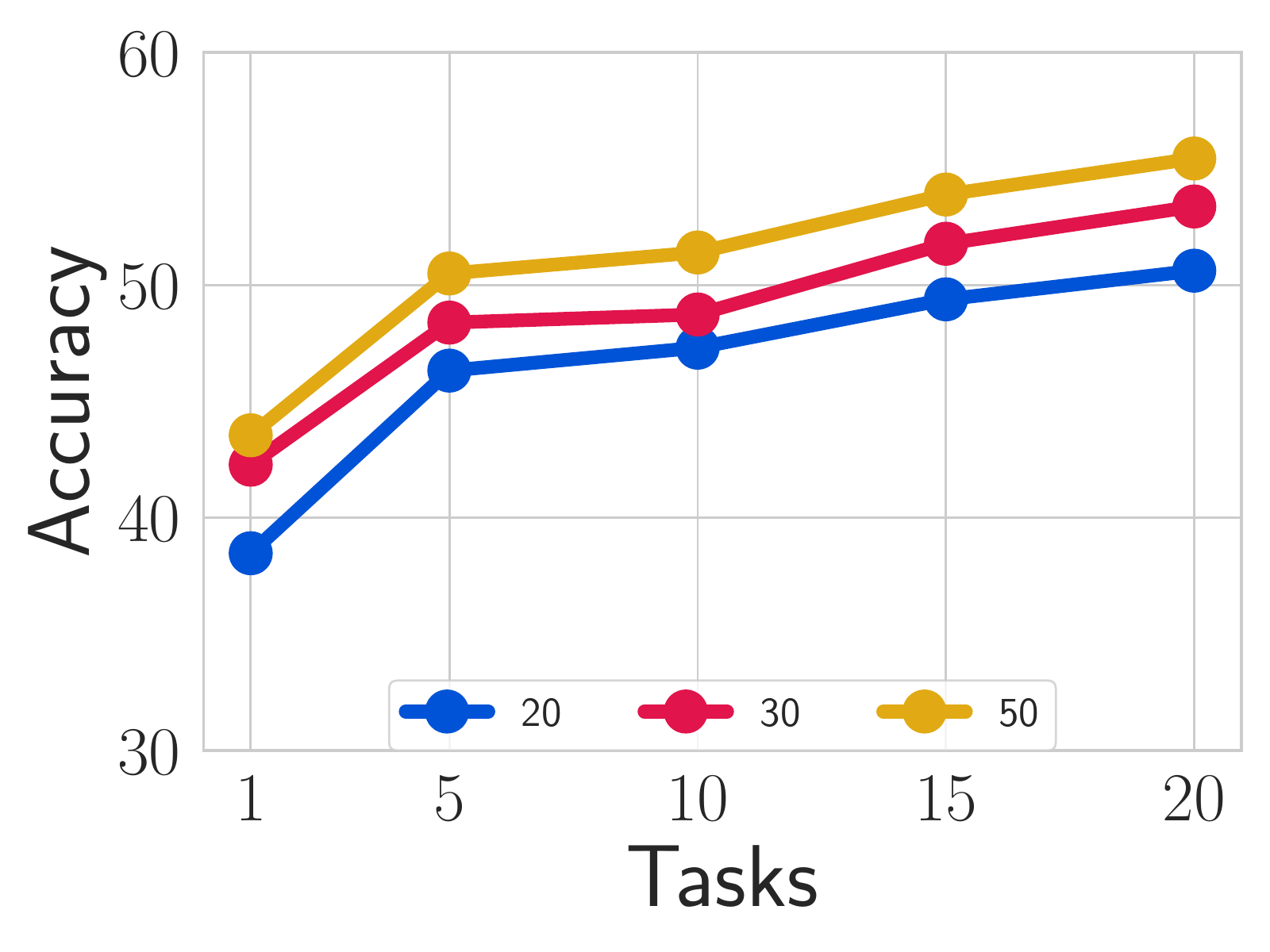}
  \caption{Split miniImageNet}
  \label{fig:positive_transfer_imagenet}
\end{subfigure}
\begin{subfigure}{.33\textwidth}
  \centering
  \includegraphics[width=.99\linewidth]{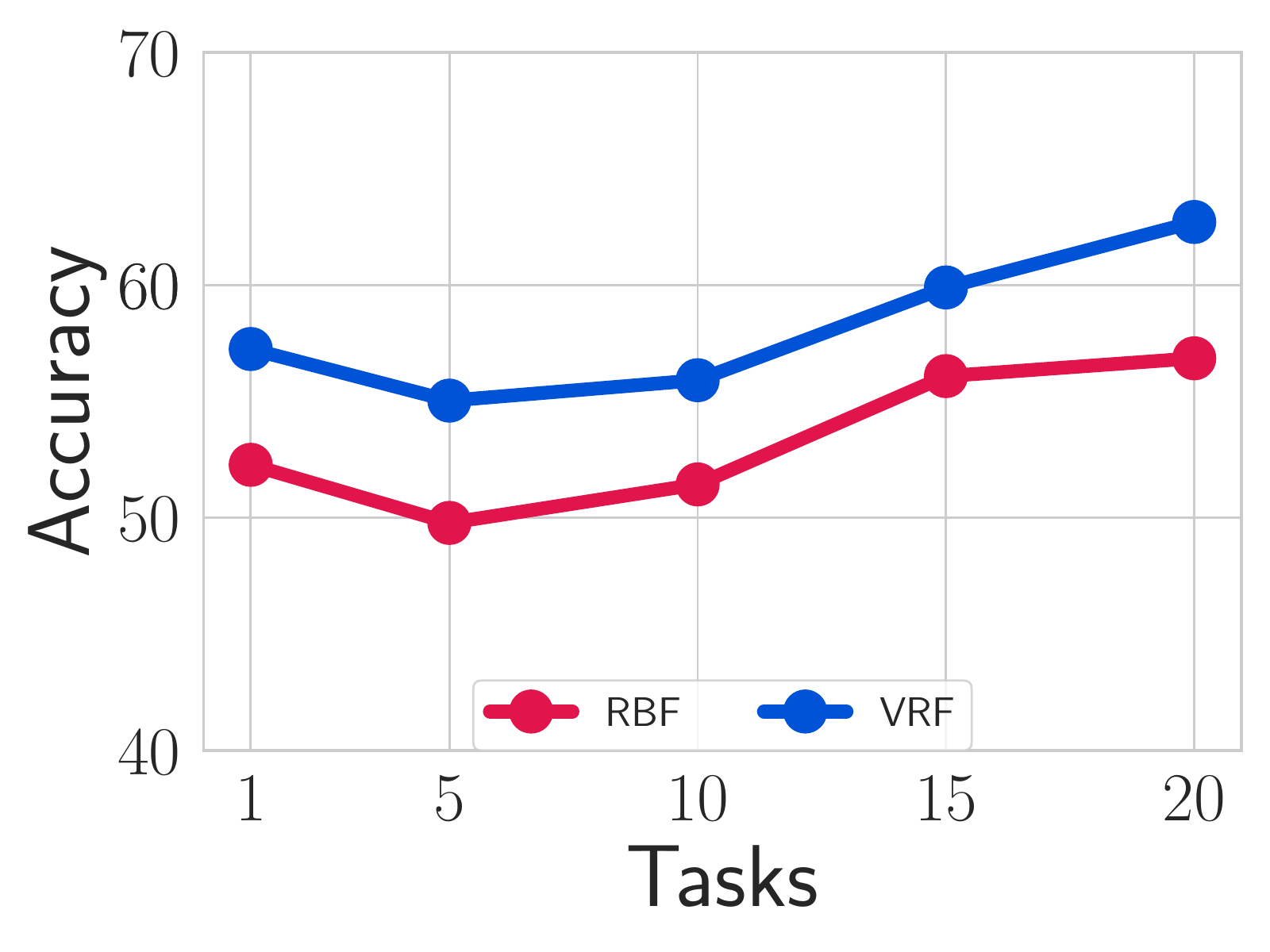}
  \caption{RBF vs. VRF on Split CIFAR100}
  \label{fig:positive_transfer_rbf_vs_vrf}
\end{subfigure}
% \vspace{-2mm}
\caption{\textbf{Influence of number of tasks}. The average accuracies of kernel continual learning by variational random features for $20$ tasks under three different coreset sizes are illustrated on Split CIFAR100 (a) and Split miniImageNet (b). Moreover, in figure (c), we show the average accuracy of two VRF and RBF kernels. As show in all figures, our proposed kernel continual learning is improved when observing more tasks.}
\label{fig:pft}
% \vspace{-5mm}
\end{figure*}

\begin{figure*}[t!]
\minipage[t]{0.32\textwidth}
  \includegraphics[width=\linewidth]{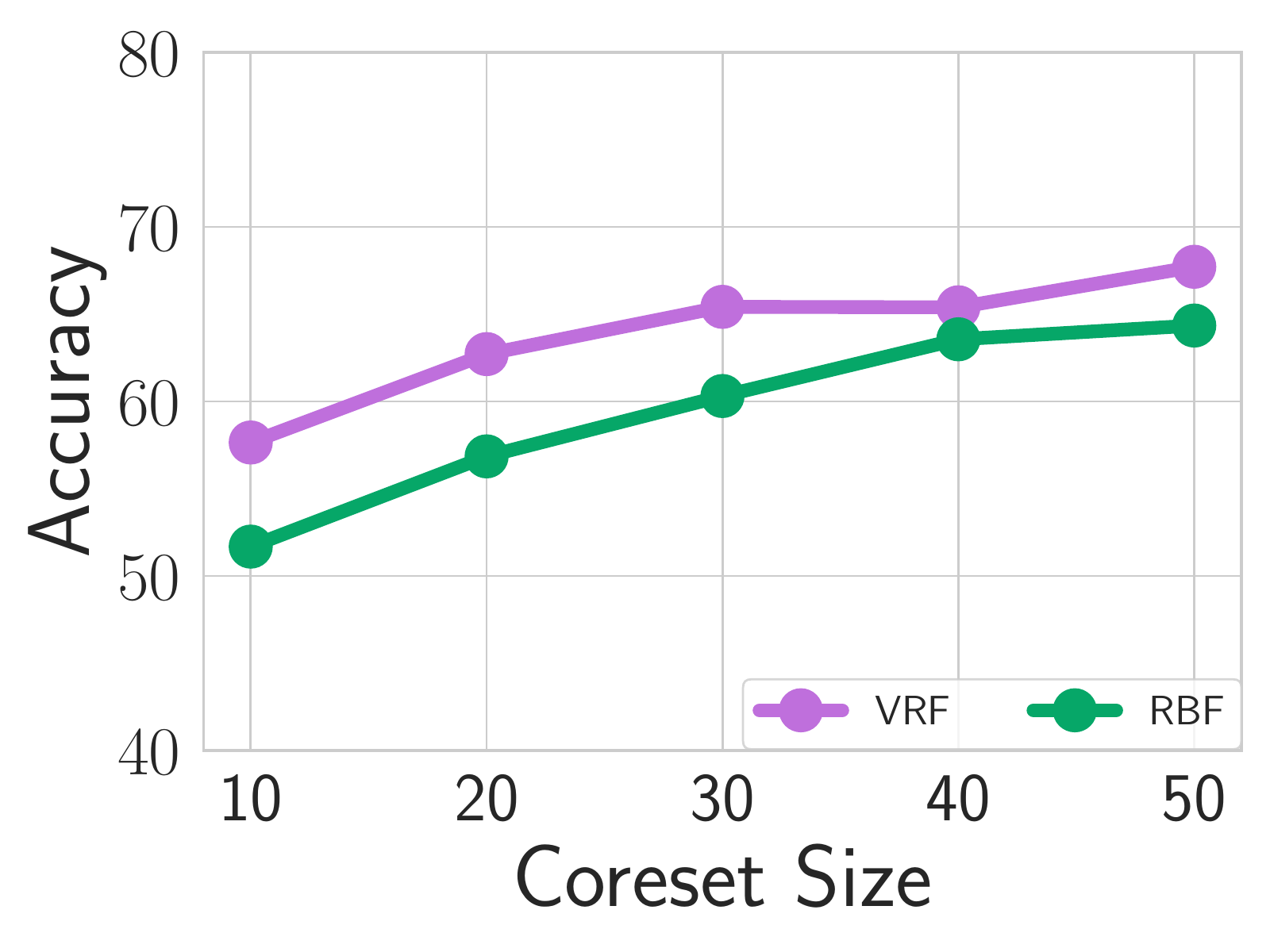}
  \caption{\textbf{Memory benefit of Variational Random Features}. To achieve similar performance, variational random features need a smaller coreset size compared to RBF kernels, showing the benefit of variational random features for kernel continual learning.} 
  \label{fig:rbf-vs-rff}
\endminipage\hfill
\minipage[t]{0.32\textwidth}
  \includegraphics[width=\linewidth]{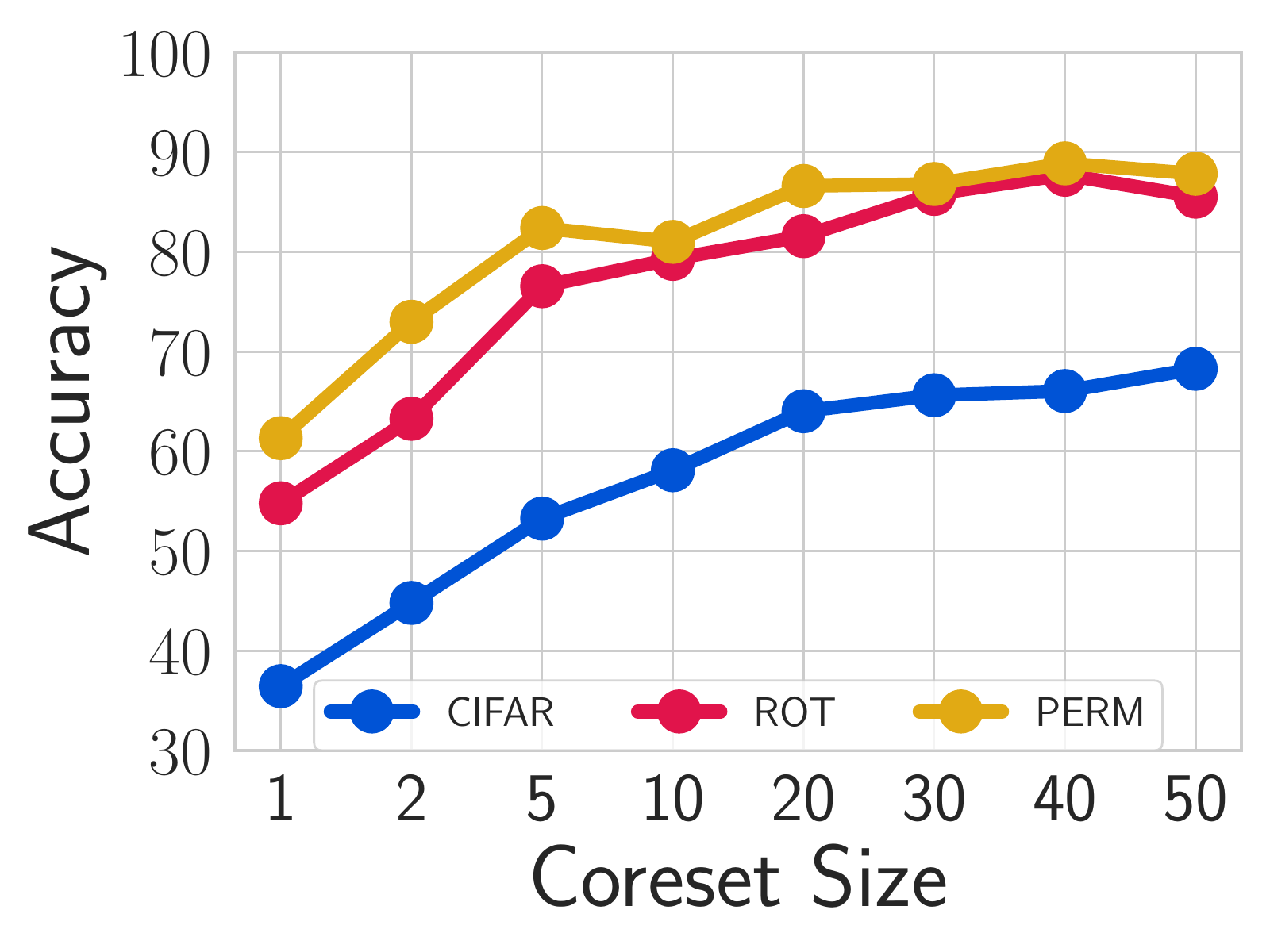}
  \caption{\textbf{How much inference memory?} Enlarging the coreset size of the VRF kernel leads to improvement in performance on all datasets. The more challenging the dataset, the more memory size helps.}
  %, we recommend a size of 30 for miniImageNet and 20 for the other three datasets. }
  \label{fig:core_vs_acc}
\endminipage\hfill
\minipage[t]{0.32\textwidth}%
  \includegraphics[width=\linewidth]{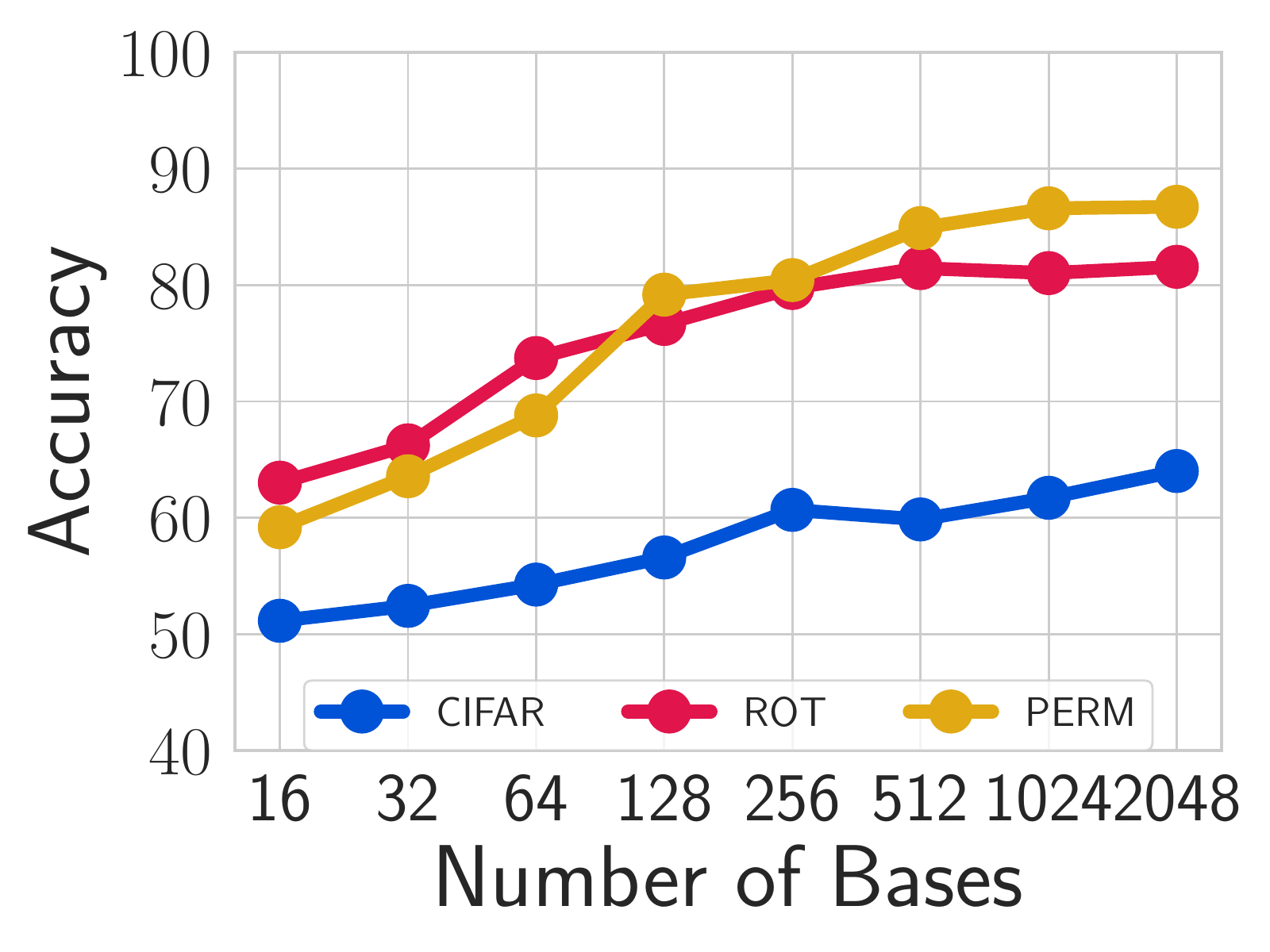}
  \caption{\textbf{How many Random Bases?} In general, a larger number of random Fourier bases consistently improves the performance on all benchmarks. With the relatively small number of $256$ bases, our variational random features can deliver good performance. % \cs{Can we report miniImageNet here as well?}
  }
  \label{fig:basis_vs_acc}
\endminipage
\end{figure*}

\begin{table}[h]
\centering
%\vspace{-2mm}
\caption{\textbf{How much inference memory?} Increasing the coreset size has only a minimal impact on time complexity at inference.}
 \vspace{2mm}
\label{tab:time_coreset}
\resizebox{.8\columnwidth}{!}{%
\begin{tabular}{@{}lcccc@{}}
\toprule
& \multicolumn{4}{c}{\textbf{Split CIFAR100}}\\
\cmidrule(lr){2-5}
%Coreset Size 
& 5 & 10 & 20 & 40 \\ 
\midrule
Time (s) & 0.0017 & 0.0017 & 0.0017 & 0.0018 \\
\bottomrule
\end{tabular}%
}
\end{table}

\begin{table}[h]
\centering
%\vspace{-2mm}
\caption{\textbf{How may Random Bases?} Increasing the number of Random Bases leads to an increased time complexity at inference.}
\vspace{2mm}
\label{tab:time_random_bases}
\resizebox{.8\columnwidth}{!}{%
\begin{tabular}{@{}lcccc@{}}
\toprule
& \multicolumn{4}{c}{\textbf{Split CIFAR100}}\\
\cmidrule(lr){2-5}
 %Bases 
 & 256 & 512 & 1024 & 2048 \\ 
\midrule
Time (s) & 0.0014 & 0.0015 & 0.0015 & 0.0017 \\
\bottomrule
\end{tabular}%
}
\vspace{-2mm}
\end{table}

\paragraph{Influence of Number of Tasks} Next we ablate the robustness of our proposal when the number of tasks increase. We report results with three different coreset sizes on Split CIFAR100 and Split miniImageNet
in Figure~\ref{fig:pft} (a) and (b). As can be seen, our method achieves increasingly better performance as the number of tasks increases, indicating that knowledge is transferred forward from previous tasks to future tasks. 
The observed positive transfer is likely due to the shared parameters in the feature extractors and amortization networks, as they allow knowledge to be transferred across tasks. 
We again show a comparison between variational random features and a predefined RBF kernel in Figure~\ref{fig:pft} (c). The performance for variational random features increases faster than the RBF kernel when observing more tasks. This might be due to the amortization network shared among tasks, which enables knowledge to be transferred across tasks as well, indicating the benefit of learning data-driven kernels by our variational random features.

%\paragraph{GIVE ABLATION A NAME}
%\cs{This does not belong to the kernel comparison right? Where does it belong?} \xz{This paragraph is talking about Figure 2, I am not sure if this figure is really needed} \cs{me neither}
%To gain insights, in Figure~\ref{fig:5-run-5-task-acc} we show the performance changes along with learning the first five tasks. We observe that accuracy have a subtle and slight decaying trend on Rotated and Permuted MINIST datasets. However, on the Split CIFAR dataset, it is shown that at beginning of the training of each task, the accuracy starts from mediocre values, and then it decays. But the accuracy on each previous task improves again along with observing new tasks.
% \cs{I miss a connection with the next section, which continues the discussion on kernels, but it is not clear to the reader. Maybe remove the bold paragraph heading and refer to the `VRF kernel' instead of `our data-driven model'.}

%

\begin{figure*}[h]
% \vspace{-3mm}
\centering
\begin{subfigure}{.33\textwidth}
  \centering
  \includegraphics[width=.99\linewidth]{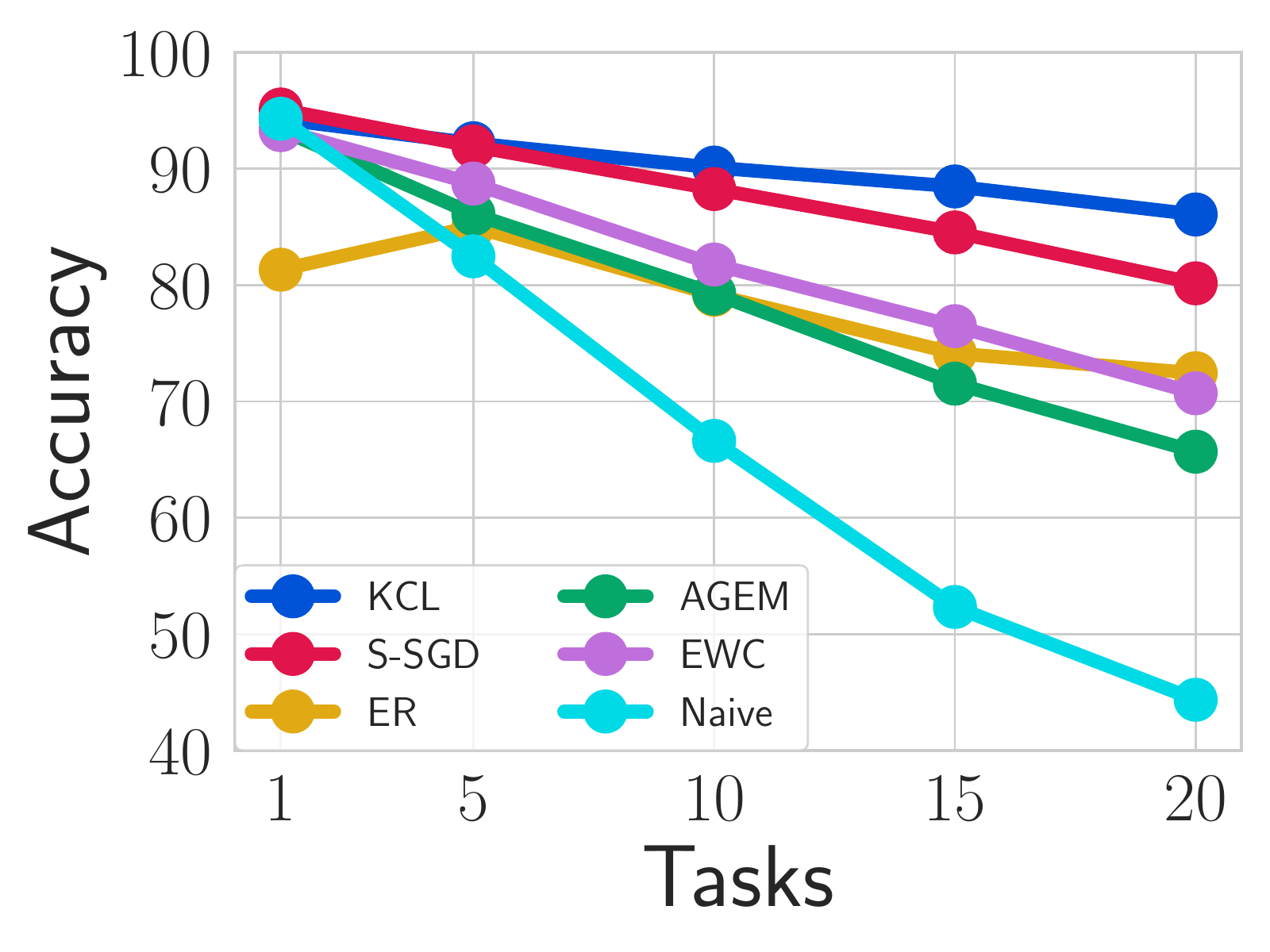}
  \caption{Permuted MNIST}
  \label{fig:5-run-20-tasks-avg-acc-pm}
\end{subfigure}%
\begin{subfigure}{.33\textwidth}
  \centering
  \includegraphics[width=.99\linewidth]{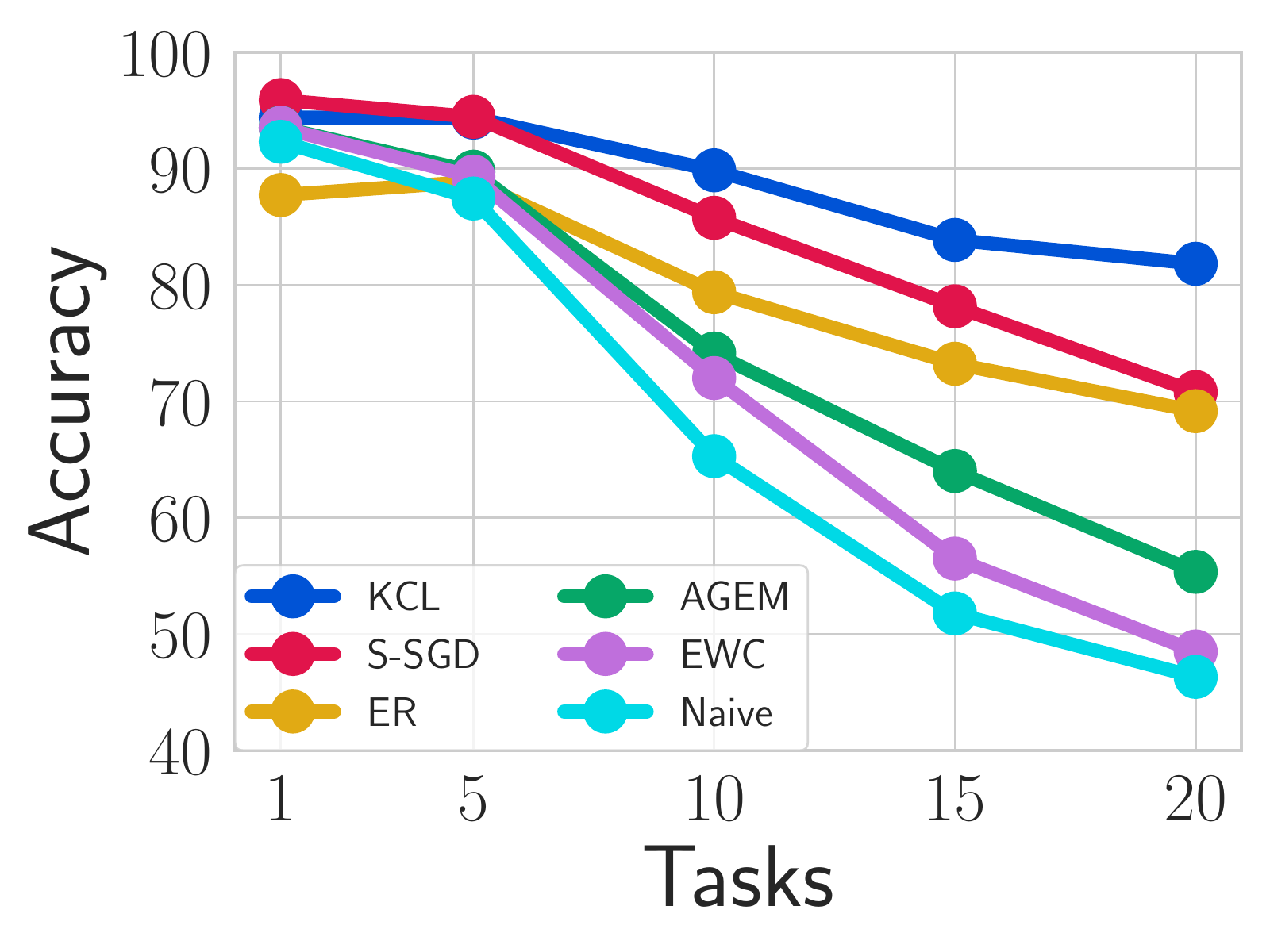}
  \caption{Rotated MNIST}
  \label{fig:5-run-20-tasks-avg-acc-rm}
\end{subfigure}
\begin{subfigure}{.33\textwidth}
  \centering
  \includegraphics[width=.99\linewidth]{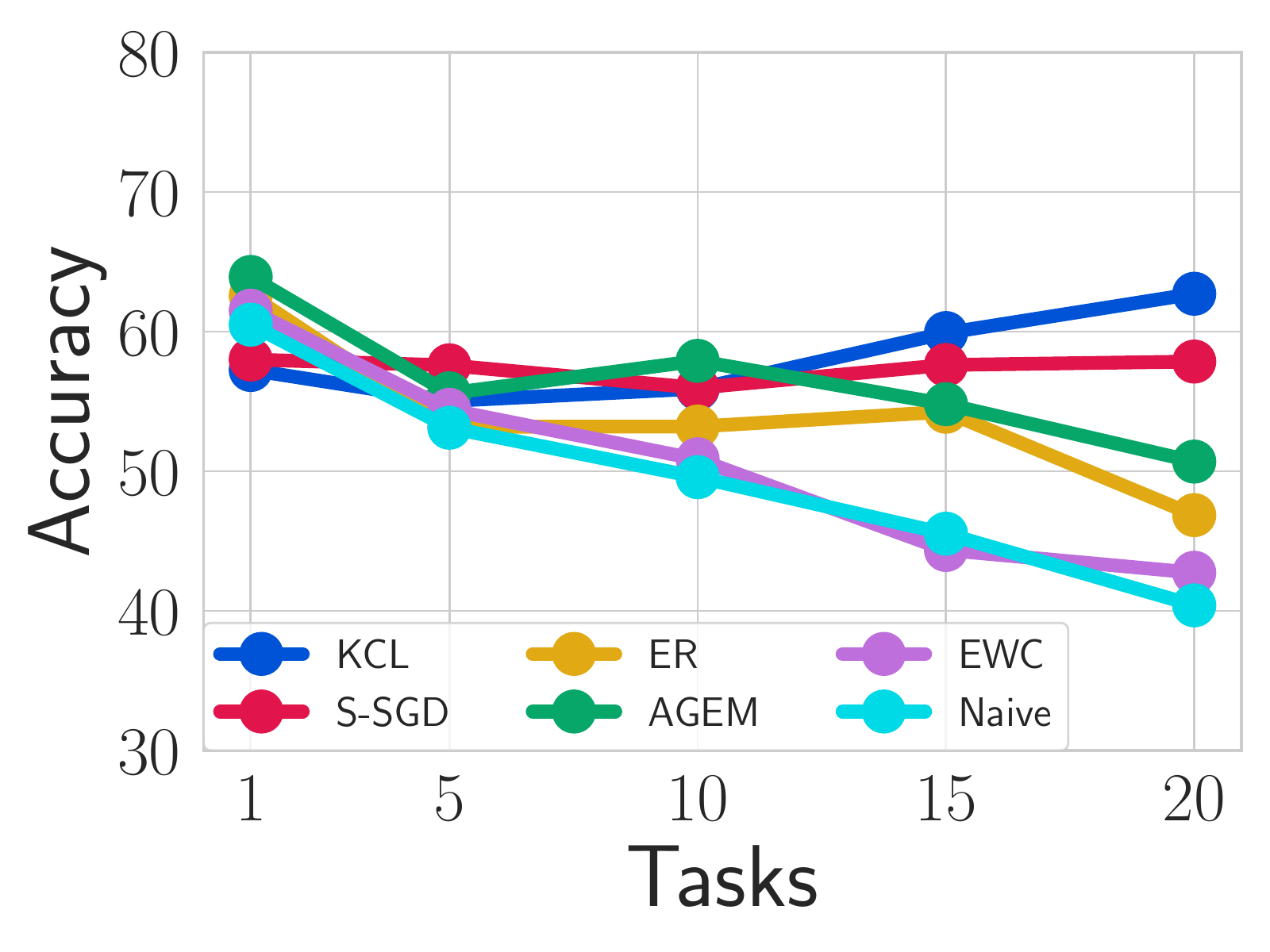}
  \caption{Split CIFAR100}
  \label{fig:5-run-20-tasks-avg-acc-cifar}
\end{subfigure}
% \vspace{-2mm}
\caption{
\textbf{Comparison between the state of the art} and our
kernel continual learning by variational random features over 20 consecutive tasks, in terms of average accuracy. Our model consistently performs better than other methods with less accuracy drop on Rotated and Permuted MNIST and, further, the performance even starts to increase when observing more tasks on the challenging Split CIFAR100 dataset.} 
\label{fig:ablation}
\vspace{-2mm}
\end{figure*}

%\cs{Is this paragraph needed here? It does not make sense to me:}
%The variational random features share the amortized inference network across tasks for generating random Fourier basis. This alleviates direct task interference in contrast to sharing classifiers
%\cs{Do not understand what you want to say, or why working as a hypernetwork is beneficial.}
%More importantly, the amortized inference network work as a hypernetwork, which allows positive knowledge transfer among tasks in a higher level instead of the classifier level~\cite{von2019continual}. 

%%experiments
% To assess the benefit of an adaptive data-driven kernel in continual learning we perform an ablation where we replace the RFF kernel with a simple radial basis function (RBF) kernel. \cs{on what dataset?} Figure~\ref{fig:ablation-rbf-rff} shows a simple kernel learning method such as RBF  outperforms simple methods such~\cite{EWC, AGEM} in terms of accuracy \cs{not visible in figure!}. We can further improve over a fixed pre-defined RBF kernel, by obtaining each task kernel in an adaptive fashion. \cs{what is the compute cost compared to RBF? Is it worth it?}

%\cs{Pascal asked a good question about training time, it is fair to demonstrate/discuss that downside also during ablation. Maybe it is not that bad?} \mo{I have checked this part and I have found there is not any difference between running time of different kernels.}

\begin{table*}[h]
\centering
\caption{\textbf{Comparison to the state-of-the-art.} Results for other methods on Permuted MNIST, Rotated MNIST and Split CIFAR100 are adopted from~\citet{mirzadeh2020understanding}. For Split miniImageNet results are from \citet{chaudhry2020continual}.We include columns denoting \textit{if} and \textit{when} memory is used. In all cases, kernel continual learning is best.}
\vspace{2mm}
\resizebox{1\linewidth}{!}{%
\begin{tabular}{lccccccccccc}
\toprule
\multirow{2}{*}{\textbf{Method}} & 
\multicolumn{2}{c}{\textbf{Memory}}
& \multicolumn{2}{c}{\textbf{Permuted MNIST}} & \multicolumn{2}{c}{\textbf{Rotated MNIST}} & \multicolumn{2}{c}{\textbf{Split CIFAR100}} & \multicolumn{2}{c}{\textbf{Split miniImageNet}} \\ \cmidrule(lr){2-3} \cmidrule(lr){4-5} \cmidrule(lr){6-7} \cmidrule(lr){8-9} \cmidrule(lr){10-11}
 & If & When & Accuracy & Forgetting & Accuracy & Forgetting & Accuracy & Forgetting & Accuracy & Forgetting\\ \midrule
Lower Bound: Naive-SGD \cite{mirzadeh2020understanding} & \xmark & - & 44.4\scriptsize{$\pm$2.46} & 0.53\scriptsize{$\pm$0.03} & 46.3\scriptsize{$\pm$1.37} & 0.52\scriptsize{$\pm$0.01} & 40.4\scriptsize{ $\pm$2.83} & 0.31\scriptsize{$\pm$0.02} & 36.1\scriptsize{$\pm$ 1.31} & 0.24\scriptsize{$\pm$ 0.03} \\
EWC \cite{kirkpatrick2017overcoming} & \xmark & -  & 70.7\scriptsize{$\pm$1.74} & 0.23\scriptsize{$\pm$0.01} & 48.5\scriptsize{$\pm$1.24} & 0.48\scriptsize{$\pm$0.01} & 42.7\scriptsize{$\pm$1.89} & 0.28\scriptsize{$\pm$0.03} & 34.8\scriptsize{$\pm$ 2.34} & 0.24\scriptsize{ $\pm$ 0.04} \\
%AGEM 
AGEM \cite{AGEM} & \cmark & Train  & 65.7\scriptsize{$\pm$0.51} & 0.29\scriptsize{$\pm$0.01} & 55.3\scriptsize{$\pm$1.47} & 0.42\scriptsize{$\pm$0.01} & 50.7\scriptsize{$\pm$2.32} & 0.19\scriptsize{$\pm$0.04} & 42.3\scriptsize{$\pm$ 1.42} & 0.17\scriptsize{$\pm$ 0.01}\\
ER-Reservoir \cite{Chaudhry2019OnTE} & \cmark & Train   & 72.4\scriptsize{$\pm$0.42} & 0.16\scriptsize{$\pm$0.01} & 69.2\scriptsize{$\pm$1.10} & 0.21\scriptsize{$\pm$0.01} & 46.9\scriptsize{$\pm$0.76} & 0.21\scriptsize{$\pm$0.03} & 49.8\scriptsize{$\pm$ 2.92} & 0.12\scriptsize{$\pm$ 0.01}\\
Stable SGD \cite{mirzadeh2020understanding} & \xmark & -  &  80.1\scriptsize{$\pm$0.51} & 0.09\scriptsize{$\pm$0.01} & 70.8\scriptsize{$\pm$0.78} & 0.10\scriptsize{$\pm$0.02} & 59.9\scriptsize{$\pm$1.81} & 0.08\scriptsize{$\pm$0.01} & - & - \\
\textit{\textbf{Kernel Continual Learning}} & \cmark & \textbf{Test}  & \textbf{85.5\scriptsize{$\pm$0.78}} & \textbf{0.02\scriptsize{$\pm$0.00}} & \textbf{81.8\scriptsize{$\pm$0.60}} & \textbf{0.01\scriptsize{$\pm$0.00}} & \textbf{62.7\scriptsize{$\pm$0.89}} & \textbf{0.06\scriptsize{$\pm$0.01}}  & \textbf{53.3\scriptsize{$\pm$ 0.57}} & \textbf{0.04\scriptsize{$\pm$ 0.00}}\\
\midrule
Upper Bound: multi-task learning \cite{mirzadeh2020understanding}  & \xmark & - & 86.5\scriptsize{$\pm$0.21} & 0.0 & 87.3\scriptsize{$\pm$0.47} & 0.0 & 64.8\scriptsize{$\pm$0.72} & 0.0 & 65.1 & 0.0\\
\bottomrule
\end{tabular}%
%\footnotesize{$^{\dagger}$~.Results for miniImageNet adopted from \cite{chaudhry2020continual}}
%
}
\label{tab:compare-20tasks}
\vspace{-2mm}
\end{table*}

% \begin{table}[h]
% \centering
% \vspace{-2mm}
% \caption{Coreset size \textit{vs.} time complexity}
% % \vspace{-2mm}
% \label{tab:time_coreset}
% \resizebox{.75\columnwidth}{!}{%
% \begin{tabular}{@{}lcccc@{}}
% \toprule
% Coreset Size & 5 & 10 & 20 & 40 \\ 
% \midrule
% Time (s) & 0.0017 & 0.0017 & 0.0017 & 0.0018 \\
% \bottomrule
% \end{tabular}%
% }
% \end{table}

% \begin{table}[h]
% \centering
% \vspace{-2mm}
% \caption{Random bases \textit{vs.} time complexity}
% % \vspace{-2mm}
% \label{tab:time_random_bases}
% \resizebox{.75\columnwidth}{!}{%
% \begin{tabular}{@{}lcccc@{}}
% \toprule
%  Bases & 256 & 512 & 1024 & 2048 \\ 
% \midrule
% Time (s) & 0.0014 & 0.0015 & 0.0015 & 0.0017 \\
% \bottomrule
% \end{tabular}%
% }
% \vspace{-2mm}
% \end{table}

\paragraph{Memory benefit of Variational Random Features}
To further demonstrate the memory benefit of data-driven kernel learning, we compare variational random features with a predefined RBF kernel in Figure~\ref{fig:rbf-vs-rff}. We consider five different coreset sizes. Variational random features exceed the RBF kernel consistently. For instance, With a smaller coreset size of 20, variational random features can achieve similar performance to the RBF kernel with a larger coreset size of 40. This demonstrates that learning task-specific kernels in a data driven way enables us to use less memory than with a pre-defined kernel.

\paragraph{How much inference memory?}
Since kernel continual learning does not need to replay and only uses memory for inference, the coreset size plays a crucial role. We therefore ablate its influence on Rotated MNIST, Permuted MNIST, and Split CIFAR100 
%and miniImageNet 
by varying the coreset size as 1, 2, 5, 10, 20, 30, 40, and 50. Here, the number of random bases is set to 1024 for Rotated MNIST and Permuted MNIST, and 2048 for Split CIFAR100. %as well as Split miniImageNet. %\cs{what setting for miniImageNet?}. %\cs{why?}
The results in Figure~\ref{fig:core_vs_acc} show that increasing the coreset size from 1 to 5 results in a steep accuracy increase for all datasets. This continues depending on the difficulty of the dataset. For Split CIFAR100, the results start to saturate after a coreset size of 20.
% and for miniImageNet the plateau starts at 30. 
This is expected as increasing the number of samples in a coreset allows us to better infer the random Fourier bases with more data from the task, therefore resulting in more representative and descriptive kernels. In the remaining experiments we use a coreset size of $20$ for Rotated MNIST, Permuted MNIST and Split CIFAR100, and a coreset size of $30$ for miniImageNet (see supplementary materials). We also ablate the effect of the coreset size on time complexity in Table \ref{tab:time_coreset}. Indeed, it shows that increasing the coreset size only comes with a limited cost increase at inference time.
%
% \cs{I don't understand what point you try to make with this statement:}
% As a result, our proposed method equips with an adjusting knob to control trade-off between the accuracy vs. storage overhead by changing the coreset size. \cs{So? What value do you use for the remaining experiment?}

\paragraph{How many Random Bases?} 
When approximating VRF kernels the number of random Fourier bases is a important hyperparameter. In principle, a larger number of random Fourier bases should yield a better approximation of kernels, leading to better classification accuracy. Here, we investigate its effect on the continual learning accuracy. Results with different numbers of bases are shown in Figure~\ref{fig:basis_vs_acc} on  Rotated MNIST, Permutated MINST and Split CIFAR100. As expected, the performance increases with a larger number of random Fourier bases, but with a relatively small number of $256$ bases, our method already performs well on all datasets. Table \ref{tab:time_random_bases} further shows the impact of the number of random bases on time complexity. It highlights that increasing the number of random bases comes with an increasing computation time for the model at inference time.

\paragraph{Comparison to the state-of-the-art}

We compare kernel continual learning with alternative methods on the four benchmarks. The accuracy and forgetting scores in Table~\ref{tab:compare-20tasks} for Rotated MNIST, Permuted MNIST and Split CIFAR100 are all adopted from~\cite{mirzadeh2020understanding}, and results for miniImageNet are from \cite{chaudhry2020continual}. The ``\textit{if}'' column indicates whether a model utilizes memory and if so, the ``\textit{when}'' column denotes whether the memory data are used during training time or test time. Our method achieves better performance in terms of average accuracy and average forgetting. Moreover, compared to memory-based methods such as A-GEM \cite{AGEM} and ER-Reservoir \cite{Chaudhry2019OnTE}, which replay over previous tasks (\textit{when} $=$ Train), kernel continual learning does not require replay, enabling our method to be efficient during training time.  Further, for the most challenging miniImageNet dataset, kernel continual learning also performs better than other methods, both in terms of accuracy and forgetting. In Figure~\ref{fig:ablation}, we compare our kernel continual learning by variational random features with other methods in terms of average accuracy over 20 consecutive tasks. Our method performs consistently better. It is worth noting that on the relatively challenging Split CIFAR100 dataset, the accuracy of our method drops a bit at the beginning but starts to increase when observing more tasks. This indicates a positive forward transfer from previous tasks to future tasks. All hyperparameters for reproducing the results in Figure~\ref{fig:ablation} and Table~\ref{tab:compare-20tasks} are provided in the supplementary materials.

% For providing a good comparison, we provide several methods that address continual learning in a different manner. For instance, EWC is regularization-based, and A-GEM and ER-Reservoir refer to replay-based model. Stable SGD is another simple yet effective continual learning which is provided in~\cite{mirzadeh2020understanding}.
% \cs{What is the conclusion, why does it work better?}

% \cs{I don't understand what point you are trying to make here.}
% In table~\ref{tab:compare-20tasks-memory-extended}, we extend the memory size of replay-based approaches such as AGEM and ER-Reservoir. This memory size is exactly the same as the coreset size we use for each dataset. To reproduce results, using code base~\footnote{https://git.io/JtkfC}, we increase the memory size and retrain the model. This table still shows that even in the case that we increase the memory size of other approaches, our methods performance is still competitive. 

% \section{Discussion}
% Here we need to talk about positive backward transfer.
% \begin{figure*}[t!]
% \centering
% \includegraphics[width=.9\textwidth]{figs/cifar_20_tasks.pdf}
% % \usepackage{amsthm}
% \caption{Positive Backward Transfer}
% \label{fig:PBT}
% \end{figure*}

%%conclusion
\section{Conclusion}
In this paper, we introduce kernel continual learning, a simple but effective variation of continual learning with kernel-based classifiers. To mitigate catastrophic forgetting, instead of using shared classifiers across tasks, we propose to learn task-specific classifiers based on kernel ridge regression. Specifically, we deploy an episodic memory unit to store a subset of training samples for each task, which is referred to as the coreset. We formulate kernel learning as a variational inference problem by treating random Fourier bases as the latent variable to be inferred from the coreset. By doing so, we are able to generate an adaptive kernel for each task while requiring a relatively small memory size. We conduct extensive experiments on four benchmark datasets for continual learning. Our thorough ablation studies demonstrate the effectiveness of kernels for continual learning and the benefits of variational random features in learning data-driven kernels for continual learning. Our kernel continual learning already achieves state-of-the-art performance on all benchmarks, while opening up many other possible connections between kernel methods and continual learning. 

%\cs{The references are still sloppy. Do not give the reviewers the impression you are sloppy. Easy to fix.}

%\cs{The references are a bit of a mess, pls cleanup and use consistent names for same venues. Avoid unnecessary clutter. Use correct bibtype. Supplement missing venues. What is the general ICML syle?}

% In the unusual situation where you want a paper to appear in the
% references without citing it in the main text, use \nocite

\section*{Acknowledgements}
This work is financially supported by the Inception Institute of Artificial Intelligence, the University of Amsterdam and the allowance Top consortia for Knowledge and Innovation (TKIs) from the Netherlands Ministry of Economic Affairs and Climate Policy.

\bibliography{kcl}

\begin{thebibliography}{58}
\providecommand{\natexlab}[1]{#1}
\providecommand{\url}[1]{\texttt{#1}}
\expandafter\ifx\csname urlstyle\endcsname\relax
  \providecommand{\doi}[1]{doi: #1}\else
  \providecommand{\doi}{doi: \begingroup \urlstyle{rm}\Url}\fi

\bibitem[Aljundi et~al.(2018)Aljundi, Babiloni, Elhoseiny, Rohrbach, and
  Tuytelaars]{MAS}
Aljundi, R., Babiloni, F., Elhoseiny, M., Rohrbach, M., and Tuytelaars, T.
\newblock Memory aware synapses: Learning what (not) to forget.
\newblock In \emph{European Conference on Computer Vision}, 2018.

\bibitem[Bach et~al.(2004)Bach, Lanckriet, and Jordan]{bach2004multiple}
Bach, F.~R., Lanckriet, G.~R., and Jordan, M.~I.
\newblock Multiple kernel learning, conic duality, and the {SMO} algorithm.
\newblock In \emph{International Conference on Machine Learning}, 2004.

\bibitem[Carratino et~al.(2018)Carratino, Rudi, and
  Rosasco]{carratino2018learning}
Carratino, L., Rudi, A., and Rosasco, L.
\newblock Learning with sgd and random features.
\newblock In \emph{Advances in Neural Information Processing Systems}, 2018.

\bibitem[Chaudhry et~al.(2018)Chaudhry, Dokania, Ajanthan, and
  Torr]{Chaudhry2018RiemannianWF}
Chaudhry, A., Dokania, P.~K., Ajanthan, T., and Torr, P. H.~S.
\newblock Riemannian walk for incremental learning: Understanding forgetting
  and intransigence.
\newblock In \emph{European Conference on Computer Vision}, 2018.

\bibitem[Chaudhry et~al.(2019{\natexlab{a}})Chaudhry, Ranzato, Rohrbach, and
  Elhoseiny]{AGEM}
Chaudhry, A., Ranzato, M., Rohrbach, M., and Elhoseiny, M.
\newblock Efficient lifelong learning with {A}-{GEM}.
\newblock In \emph{International Conference on Learning Representations},
  2019{\natexlab{a}}.

\bibitem[Chaudhry et~al.(2019{\natexlab{b}})Chaudhry, Rohrbach, Elhoseiny,
  Ajanthan, Dokania, Torr, and Ranzato]{Chaudhry2019OnTE}
Chaudhry, A., Rohrbach, M., Elhoseiny, M., Ajanthan, T., Dokania, P.~K., Torr,
  P. H.~S., and Ranzato, M.
\newblock On tiny episodic memories in continual learning.
\newblock In \emph{Advances in Neural Information Processing Systems},
  2019{\natexlab{b}}.

\bibitem[Chaudhry et~al.(2020)Chaudhry, Khan, Dokania, and
  Torr]{chaudhry2020continual}
Chaudhry, A., Khan, N., Dokania, P.~K., and Torr, P.~H.
\newblock Continual learning in low-rank orthogonal subspaces.
\newblock In \emph{Advances in Neural Information Processing System}, 2020.

\bibitem[Cristianini et~al.(2000)Cristianini, Shawe-Taylor,
  et~al.]{cristianini2000introduction}
Cristianini, N., Shawe-Taylor, J., et~al.
\newblock \emph{An introduction to support vector machines and other
  kernel-based learning methods}.
\newblock Cambridge university press, 2000.

\bibitem[Diehl \& Cauwenberghs(2003)Diehl and Cauwenberghs]{diehl2003svm}
Diehl, C.~P. and Cauwenberghs, G.
\newblock Svm incremental learning, adaptation and optimization.
\newblock In \emph{Proceedings of the International Joint Conference on Neural
  Networks}, 2003.

\bibitem[Ebrahimi et~al.(2020)Ebrahimi, Elhoseiny, Darrell, and
  Rohrbach]{ebrahimi2019uncertainty}
Ebrahimi, S., Elhoseiny, M., Darrell, T., and Rohrbach, M.
\newblock Uncertainty-guided continual learning with bayesian neural networks.
\newblock In \emph{International Conference on Learning Representations}, 2020.

\bibitem[Goodfellow et~al.(2014)Goodfellow, Mirza, Xiao, Courville, and
  Bengio]{Goodfellow2013AnEI}
Goodfellow, I.~J., Mirza, M., Xiao, D., Courville, A., and Bengio, Y.
\newblock An empirical investigation of catastrophic forgeting in gradientbased
  neural networks.
\newblock In \emph{International Conference on Learning Representations}, 2014.

\bibitem[Gordon et~al.(2019)Gordon, Bronskill, Bauer, Nowozin, and
  Turner]{gordon2018meta}
Gordon, J., Bronskill, J., Bauer, M., Nowozin, S., and Turner, R.~E.
\newblock Meta-learning probabilistic inference for prediction.
\newblock In \emph{International Conference on Learning Representations}, 2019.

\bibitem[Hadsell et~al.(2020)Hadsell, Rao, Rusu, and
  Pascanu]{hadsell2020embracing}
Hadsell, R., Rao, D., Rusu, A.~A., and Pascanu, R.
\newblock Embracing change: Continual learning in deep neural networks.
\newblock \emph{Trends in Cognitive Sciences}, 2020.

\bibitem[Jerfel et~al.(2019)Jerfel, Grant, Griffiths, and
  Heller]{Jerfel2018ReconcilingMA}
Jerfel, G., Grant, E., Griffiths, T.~L., and Heller, K.~A.
\newblock Reconciling meta-learning and continual learning with online mixtures
  of tasks.
\newblock In \emph{Advances in Neural Information Processing Systems}, 2019.

\bibitem[Kingma \& Welling(2014)Kingma and Welling]{kingma2013auto}
Kingma, D.~P. and Welling, M.
\newblock Auto-encoding variational bayes.
\newblock In \emph{International Conference on Learning Representations}, 2014.

\bibitem[Kirkpatrick et~al.(2017)Kirkpatrick, Pascanu, Rabinowitz, Veness,
  Desjardins, Rusu, Milan, Quan, Ramalho, Grabska-Barwinska, Hassabis, Clopath,
  Kumaran, and Hadsell]{kirkpatrick2017overcoming}
Kirkpatrick, J., Pascanu, R., Rabinowitz, N., Veness, J., Desjardins, G., Rusu,
  A.~A., Milan, K., Quan, J., Ramalho, T., Grabska-Barwinska, A., Hassabis, D.,
  Clopath, C., Kumaran, D., and Hadsell, R.
\newblock Overcoming catastrophic forgetting in neural networks.
\newblock \emph{Proceedings of the National Academy of Sciences}, 114\penalty0
  (13):\penalty0 3521--3526, 2017.

\bibitem[Kolouri et~al.(2019)Kolouri, Ketz, Zou, Krichmar, and
  Pilly]{kolouri2019attention}
Kolouri, S., Ketz, N., Zou, X., Krichmar, J., and Pilly, P.
\newblock Attention-based structural-plasticity.
\newblock \emph{arXiv preprint arXiv:1903.06070}, 2019.

\bibitem[Lange et~al.(2019)Lange, Aljundi, Masana, Parisot, Jia, Leonardis,
  Slabaugh, and Tuytelaars]{Lange2019ContinualLA}
Lange, M., Aljundi, R., Masana, M., Parisot, S., Jia, X., Leonardis, A.,
  Slabaugh, G.~G., and Tuytelaars, T.
\newblock Continual learning: A comparative study on how to defy forgetting in
  classification tasks.
\newblock \emph{arXiv preprint arXiv:1909.08383}, 2019.

\bibitem[LeCun et~al.(2015)LeCun, Bengio, and Hinton]{lecun2015deep}
LeCun, Y., Bengio, Y., and Hinton, G.
\newblock Deep learning.
\newblock \emph{Nature}, 2015.

\bibitem[Lee et~al.(2017)Lee, Kim, Jun, Ha, and Zhang]{lee2017overcoming}
Lee, S.-W., Kim, J.-H., Jun, J., Ha, J.-W., and Zhang, B.-T.
\newblock Overcoming catastrophic forgetting by incremental moment matching.
\newblock In \emph{Advances in Neural Information Processing Systems}, 2017.

\bibitem[Li et~al.(2019)Li, Zhou, Wu, Socher, and Xiong]{li2019learn}
Li, X., Zhou, Y., Wu, T., Socher, R., and Xiong, C.
\newblock Learn to grow: A continual structure learning framework for
  overcoming catastrophic forgetting.
\newblock In \emph{International Conference on Machine Learning}, 2019.

\bibitem[Lopez-Paz \& Ranzato(2017)Lopez-Paz and Ranzato]{lopez2017gradient}
Lopez-Paz, D. and Ranzato, M.
\newblock Gradient episodic memory for continual learning.
\newblock In \emph{Advances in Neural Information Processing Systems}, 2017.

\bibitem[Mallya \& Lazebnik(2018)Mallya and Lazebnik]{PackNet}
Mallya, A. and Lazebnik, S.
\newblock Packnet: Adding multiple tasks to a single network by iterative
  pruning.
\newblock In \emph{IEEE Conference on Computer Vision and Pattern Recognition},
  2018.

\bibitem[Masse et~al.(2018)Masse, Grant, and Freedman]{Gating}
Masse, N.~Y., Grant, G.~D., and Freedman, D.~J.
\newblock Alleviating catastrophic forgetting using context-dependent gating
  and synaptic stabilization.
\newblock \emph{Proceedings of the National Academy of Sciences}, 115\penalty0
  (44), 2018.

\bibitem[McCloskey \& Cohen(1989)McCloskey and
  Cohen]{McCloskey1989CatastrophicII}
McCloskey, M. and Cohen, N.~J.
\newblock Catastrophic interference in connectionist networks: The sequential
  learning problem.
\newblock \emph{Academic Press}, 1989.

\bibitem[Mirzadeh et~al.(2020)Mirzadeh, Farajtabar, Pascanu, and
  Ghasemzadeh]{mirzadeh2020understanding}
Mirzadeh, S.~I., Farajtabar, M., Pascanu, R., and Ghasemzadeh, H.
\newblock Understanding the role of training regimes in continual learning.
\newblock In \emph{Advances in Neural Information Processing Systems}, 2020.

\bibitem[Nguyen et~al.(2018)Nguyen, Li, Bui, and Turner]{nguyen2017variational}
Nguyen, C.~V., Li, Y., Bui, T.~D., and Turner, R.~E.
\newblock Variational continual learning.
\newblock In \emph{International Conference on Learning Representations}, 2018.

\bibitem[Parisi et~al.(2018)Parisi, Kemker, Part, Kanan, and
  Wermter]{Parisi2018ContinualLL}
Parisi, G., Kemker, R., Part, J.~L., Kanan, C., and Wermter, S.
\newblock Continual lifelong learning with neural networks: A review.
\newblock \emph{Neural Networks}, 2018.

\bibitem[Paszke et~al.(2019)Paszke, Gross, Massa, and et. al.]{Pytorch}
Paszke, A., Gross, S., Massa, F., and et. al.
\newblock Pytorch: An imperative style, high-performance deep learning library.
\newblock 2019.

\bibitem[Patacchiola et~al.(2020)Patacchiola, Turner, Crowley, O'Boyle, and
  Storkey]{patacchiola2020bayesian}
Patacchiola, M., Turner, J., Crowley, E.~J., O'Boyle, M., and Storkey, A.
\newblock Bayesian meta-learning for the few-shot setting via deep kernels.
\newblock In \emph{Advances in Neural Information Processing Systems}, 2020.

\bibitem[Pentina \& Ben-David(2015)Pentina and Ben-David]{pentina2015multi}
Pentina, A. and Ben-David, S.
\newblock Multi-task and lifelong learning of kernels.
\newblock In \emph{International Conference on Algorithmic Learning Theory},
  2015.

\bibitem[Rahimi \& Recht(2007)Rahimi and Recht]{rahimi2007random}
Rahimi, A. and Recht, B.
\newblock Random features for large-scale kernel machines.
\newblock In \emph{Advances in Neural Information Processing Systems}, 2007.

\bibitem[Ramasesh et~al.(2021)Ramasesh, Dyer, and Raghu]{ramasesh2020anatomy}
Ramasesh, V.~V., Dyer, E., and Raghu, M.
\newblock Anatomy of catastrophic forgetting: Hidden representations and task
  semantics.
\newblock \emph{International Conference on Learning Representations}, 2021.

\bibitem[Rebuffi et~al.(2017)Rebuffi, Kolesnikov, Sperl, and
  Lampert]{Rebuffi2016iCaRLIC}
Rebuffi, S.-A., Kolesnikov, A.~I., Sperl, G., and Lampert, C.~H.
\newblock {iCaRL}: Incremental classifier and representation learning.
\newblock In \emph{IEEE Conference on Computer Vision and Pattern Recognition},
  2017.

\bibitem[Riemer et~al.(2019)Riemer, Cases, Ajemian, Liu, Rish, Tu, and
  Tesauro]{riemer2018learning}
Riemer, M., Cases, I., Ajemian, R., Liu, M., Rish, I., Tu, Y., and Tesauro, G.
\newblock Learning to learn without forgetting by maximizing transfer and
  minimizing interference.
\newblock In \emph{International Conference on Learning Representations}, 2019.

\bibitem[Ring(1998)]{ring1998child}
Ring, M.~B.
\newblock Child: A first step towards continual learning.
\newblock \emph{Learning to learn}, 1998.

\bibitem[Rios \& Itti(2018)Rios and Itti]{rios2018closed}
Rios, A. and Itti, L.
\newblock Closed-loop gan for continual learning.
\newblock In \emph{International Joint Conference on Artificial Intelligence},
  2018.

\bibitem[Ritter et~al.(2018)Ritter, Botev, and Barber]{ritter2018online}
Ritter, H., Botev, A., and Barber, D.
\newblock Online structured laplace approximations for overcoming catastrophic
  forgetting.
\newblock In \emph{Advances in Neural Information Processing Systems}, 2018.

\bibitem[Rudin(1962)]{rudin1962fourier}
Rudin, W.
\newblock \emph{Fourier analysis on groups}.
\newblock Wiley Online Library, 1962.

\bibitem[Russakovsky et~al.(2015)Russakovsky, Deng, Su, Krause, Satheesh, Ma,
  Huang, Karpathy, Khosla, Bernstein, et~al.]{russakovsky2015imagenet}
Russakovsky, O., Deng, J., Su, H., Krause, J., Satheesh, S., Ma, S., Huang, Z.,
  Karpathy, A., Khosla, A., Bernstein, M., et~al.
\newblock Imagenet large scale visual recognition challenge.
\newblock \emph{International journal of computer vision}, 2015.

\bibitem[Rusu et~al.(2016)Rusu, Rabinowitz, Desjardins, Soyer, Kirkpatrick,
  Kavukcuoglu, Pascanu, and Hadsell]{rusu2016progressive}
Rusu, A.~A., Rabinowitz, N.~C., Desjardins, G., Soyer, H., Kirkpatrick, J.,
  Kavukcuoglu, K., Pascanu, R., and Hadsell, R.
\newblock Progressive neural networks.
\newblock In \emph{Advances in Neural Information Processing Systems}, 2016.

\bibitem[Schmidhuber(2015)]{schmidhuber2015deep}
Schmidhuber, J.
\newblock Deep learning in neural networks: An overview.
\newblock \emph{Neural Networks}, 2015.

\bibitem[Sch{\"o}lkopf \& Smola(2002)Sch{\"o}lkopf and
  Smola]{smola1998learning}
Sch{\"o}lkopf, B. and Smola, A.~J.
\newblock \emph{Learning with kernels}.
\newblock MIT Press, 2002.

\bibitem[Sch{\"o}lkopf et~al.(2001)Sch{\"o}lkopf, Herbrich, and
  Smola]{scholkopf2001generalized}
Sch{\"o}lkopf, B., Herbrich, R., and Smola, A.~J.
\newblock A generalized representer theorem.
\newblock In \emph{International Conference on Computational Learning Theory},
  2001.

\bibitem[Schwarz et~al.(2018)Schwarz, Czarnecki, Luketina, Grabska-Barwinska,
  Teh, Pascanu, and Hadsell]{schwarz2018progress}
Schwarz, J., Czarnecki, W., Luketina, J., Grabska-Barwinska, A., Teh, Y.~W.,
  Pascanu, R., and Hadsell, R.
\newblock Progress \& compress: A scalable framework for continual learning.
\newblock In \emph{International Conference on Machine Learning}, 2018.

\bibitem[Shin et~al.(2017)Shin, Lee, Kim, and Kim]{shin2017continual}
Shin, H., Lee, J.~K., Kim, J., and Kim, J.
\newblock Continual learning with deep generative replay.
\newblock In \emph{Advances in Neural Information Processing Systems}, 2017.

\bibitem[Sinha \& Duchi(2016)Sinha and Duchi]{sinha2016learning}
Sinha, A. and Duchi, J.~C.
\newblock Learning kernels with random features.
\newblock In \emph{Advances in Neural Information Processing Systems}, 2016.

\bibitem[Smola \& Sch{\"o}lkopf(2004)Smola and
  Sch{\"o}lkopf]{smola2004tutorial}
Smola, A.~J. and Sch{\"o}lkopf, B.
\newblock A tutorial on support vector regression.
\newblock \emph{Statistics and computing}, 2004.

\bibitem[Titsias et~al.(2020)Titsias, Schwarz, Matthews, Pascanu, and
  Teh]{titsias2019functional}
Titsias, M.~K., Schwarz, J., Matthews, A. G. d.~G., Pascanu, R., and Teh, Y.~W.
\newblock Functional regularisation for continual learning using gaussian
  processes.
\newblock In \emph{International Conference on Learning Representations}, 2020.

\bibitem[Tossou et~al.(2019)Tossou, Dura, Laviolette, Marchand, and
  Lacoste]{tossou2019adaptive}
Tossou, P., Dura, B., Laviolette, F., Marchand, M., and Lacoste, A.
\newblock Adaptive deep kernel learning.
\newblock In \emph{Advances in Neural Information Processing Systems}, 2019.

\bibitem[Vinyals et~al.(2016)Vinyals, Blundell, Lillicrap, Kavukcuoglu, and
  Wierstra]{vinyals2016matching}
Vinyals, O., Blundell, C., Lillicrap, T., Kavukcuoglu, K., and Wierstra, D.
\newblock Matching networks for one shot learning.
\newblock In \emph{Advances in Neural Information Processing Systems}, 2016.

\bibitem[Wilson et~al.(2016{\natexlab{a}})Wilson, Hu, Salakhutdinov, and
  Xing]{wilson2016deep}
Wilson, A.~G., Hu, Z., Salakhutdinov, R., and Xing, E.~P.
\newblock Deep kernel learning.
\newblock In \emph{International Conference on Artificial Intelligence and
  Statistics}, 2016{\natexlab{a}}.

\bibitem[Wilson et~al.(2016{\natexlab{b}})Wilson, Hu, Salakhutdinov, and
  Xing]{wilson2016stochastic}
Wilson, A.~G., Hu, Z., Salakhutdinov, R., and Xing, E.~P.
\newblock Stochastic variational deep kernel learning.
\newblock In \emph{Advances in Neural Information Processing Systems},
  2016{\natexlab{b}}.

\bibitem[Wortsman et~al.(2020)Wortsman, Ramanujan, Liu, Kembhavi, Rastegari,
  Yosinski, and Farhadi]{wortsman2020supermasks}
Wortsman, M., Ramanujan, V., Liu, R., Kembhavi, A., Rastegari, M., Yosinski,
  J., and Farhadi, A.
\newblock Supermasks in superposition.
\newblock In \emph{Advances in Neural Information Processing Systems}, 2020.

\bibitem[Yoon et~al.(2018)Yoon, Yang, Lee, and Hwang]{yoon2018lifelong}
Yoon, J., Yang, E., Lee, J., and Hwang, S.~J.
\newblock Lifelong learning with dynamically expandable networks.
\newblock In \emph{International Conference on Learning Representations}, 2018.

\bibitem[Zenke et~al.(2017)Zenke, Poole, and Ganguli]{zenke2017continual}
Zenke, F., Poole, B., and Ganguli, S.
\newblock Continual learning through synaptic intelligence.
\newblock In \emph{International Conference on Machine Learning}, 2017.

\bibitem[Zhang et~al.(2019)Zhang, Wang, Lim, and Feng]{zhang2019prototype}
Zhang, M., Wang, T., Lim, J.~H., and Feng, J.
\newblock Prototype reminding for continual learning.
\newblock \emph{arXiv preprint arXiv:1905.09447}, 2019.

\bibitem[Zhen et~al.(2020)Zhen, Sun, Du, Xu, Yin, Shao, and
  Snoek]{zhen2020learning}
Zhen, X., Sun, H., Du, Y., Xu, J., Yin, Y., Shao, L., and Snoek, C.
\newblock Learning to learn kernels with variational random features.
\newblock In \emph{International Conference on Machine Learning}, 2020.

\end{thebibliography}
\bibliographystyle{icml2021}

\clearpage
\onecolumn

\section{Appendix}

% \printAffiliationsAndNotice{\icmlEqualContribution} % otherwise use the standard text.
% \appendix
% \onecolumn
In Section~\ref{sec:elbo_derive}, we provide the detailed derivation of the Evidence Lower Bound (ELBO) of our variational random features for kernel continual learning. In Section~\ref{sec:model}, we illustrate our proposed model in detail, explaining each part. Moreover, for all kernels, e.g., Linear, Polynomial, and Radial Basis Function, we discuss how the main architecture is changed accordingly. In Section~\ref{sec:hparams}, all hyperparamters are listed in a table in order to reproduce the paper results. 
%Afterwards, in section \ref{sec:paper_ablation_discussion}, we clarify about how each ablation study provided in the main paper is conducted. 
Finally, in Section~\ref{sec:paper_ablation_discussion}, we include additional ablation results on miniImageNet for the size of inference memory and number of Random Bases.

\subsection{Derivation of Evidence Lower Bound}
\label{sec:elbo_derive}
Our proposed objective in equation 10 is derived as follows:
\begin{equation}
\begin{aligned}
\ln p(\mathbf{y} | \mathbf{x}, \mathcal{D}_t\backslash \mathcal{C}_t)  & = \ln \Big[\int p(\mathbf{y} | \mathbf{x}, \boldsymbol{\omega},\mathcal{D}_t\backslash \mathcal{C}_t) p(\boldsymbol{\omega} | \mathcal{D}_t\backslash \mathcal{C}_t) d\boldsymbol{\omega}\Big]\\
&= \ln \Big[ \int p(\mathbf{y} | \mathbf{x}, \boldsymbol{\omega},\mathcal{D}_t\backslash \mathcal{C}_t) p(\boldsymbol{\omega} | \mathcal{D}_t\backslash \mathcal{C}_t) \frac{q_{\phi}(\boldsymbol{\omega} | \mathcal{C}_t)}{q_{\phi}(\boldsymbol{\omega} | \mathcal{C}_t)} d\boldsymbol{\omega} \Big]\\
&=\ln \Big[\mathbb{E}_{q_{\phi}(\boldsymbol{\omega} | \mathcal{C}_t)}{\frac{p(\mathbf{y} | \mathbf{x}, \boldsymbol{\omega},\mathcal{D}_t\backslash \mathcal{C}_t) p(\boldsymbol{\omega} | \mathcal{D}_t\backslash \mathcal{C}_t)}{q_{\phi}(\boldsymbol{\omega} | \mathcal{C}_t)}}\Big].\\
\end{aligned}
\end{equation}
By applying Jensen's inequality, we have
\begin{equation}
\begin{aligned}
\ln p(\mathbf{y} | \mathbf{x}, \mathcal{D}_t\backslash \mathcal{C}_t) &\geq \mathbb{E}_{q_{\phi} (\boldsymbol{\omega} | \mathcal{C}_t)} \Big[ \ln {\frac{p(\mathbf{y} | \mathbf{x}, \boldsymbol{\omega},\mathcal{D}_t\backslash \mathcal{C}_t) p(\boldsymbol{\omega} | \mathcal{D}_t\backslash \mathcal{C}_t)}{q_{\phi}(\boldsymbol{\omega} | \mathcal{C}_t)}}\Big]\\
&= \mathbb{E}_{q_{\phi} (\boldsymbol{\omega} | \mathcal{C}_t)} \Big[\ln {p(\mathbf{y} | \mathbf{x}, \boldsymbol{\omega},\mathcal{D}_t\backslash \mathcal{C}_t)} + \ln {\frac{p(\boldsymbol{\omega} | \mathcal{D}_t\backslash \mathcal{C}_t)}{q_{\phi}(\boldsymbol{\omega} | \mathcal{C}_t)}}\Big]\\
&= \mathbb{E}_{q_{\phi} (\boldsymbol{\omega} | \mathcal{C}_t)} \Big[\ln {p(\mathbf{y} | \mathbf{x}, \boldsymbol{\omega},\mathcal{D}_t\backslash \mathcal{C}_t)} \Big] + \mathbb{E}_{q_{\phi} (\boldsymbol{\omega} | \mathcal{C}_t)} \Big[ \ln {\frac{p(\boldsymbol{\omega} | \mathcal{D}_t\backslash \mathcal{C}_t)}{q_{\phi}(\boldsymbol{\omega} | \mathcal{C}_t)}}\Big]\\
&= \mathbb{E}_{q_{\phi} (\boldsymbol{\omega} | \mathcal{C}_t)} \Big[\ln {p(\mathbf{y} | \mathbf{x}, \boldsymbol{\omega},\mathcal{D}_t\backslash \mathcal{C}_t)} \Big] - D_{\mathrm{KL}}\big[q_{\phi}(\boldsymbol{\omega} | \mathcal{C}_t) \| p_{\gamma}(\boldsymbol{\omega} | \mathcal{D}_t\backslash \mathcal{C}_t)\big]\\
&=\mathcal{L}_{\mathrm{ELBO}}.
\label{elbo}
\end{aligned}
\end{equation}

\subsection{Model Details}
\label{sec:model}
\begin{figure*}[!htp]
\centering
\includegraphics[width=.9\textwidth]{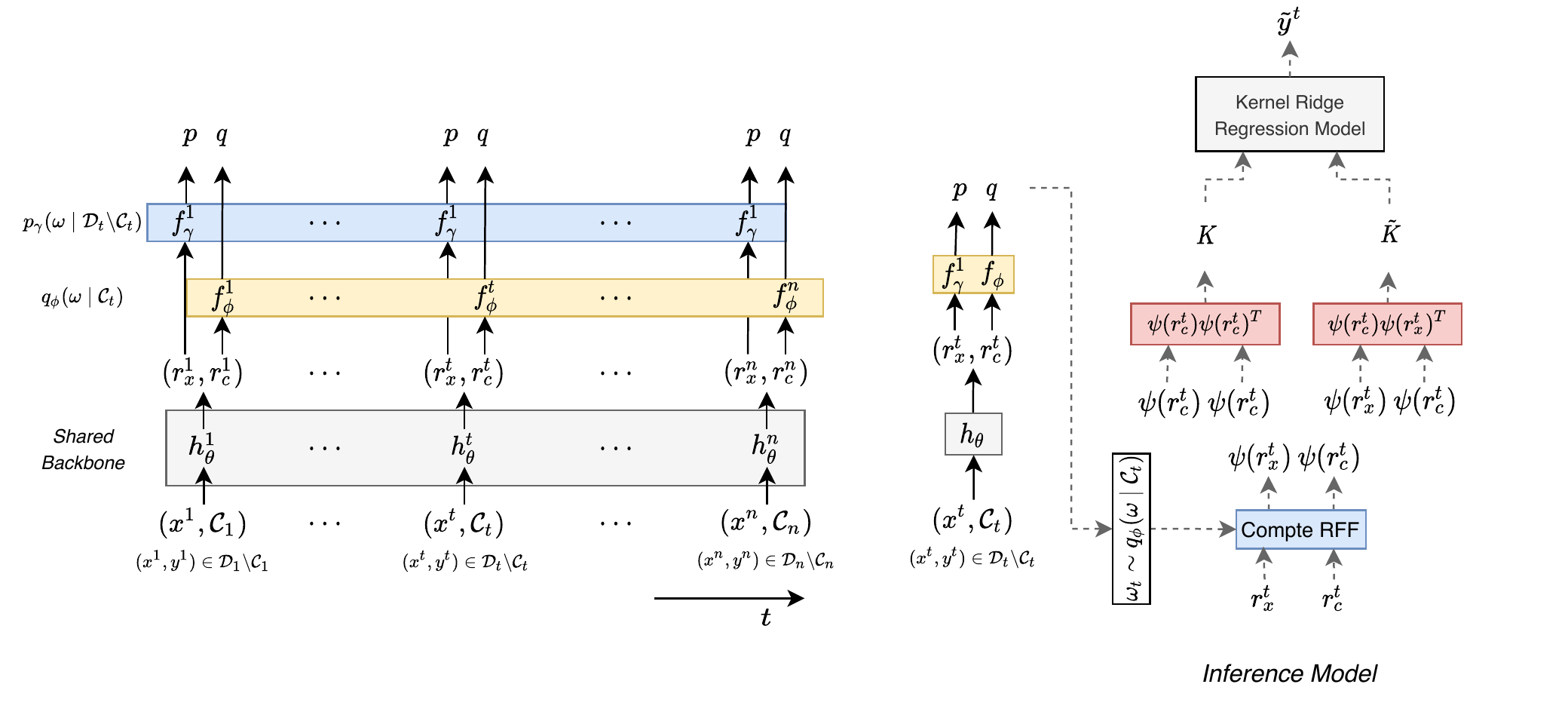}
\caption{Kernel continual learning model with variational random features.}
\label{fig:model}
\end{figure*}

We provide the computational graph of our kernel continual learning with variational random features in Figure \ref{fig:model}. Our method consists of three networks. $h_\theta$ is the backbone network shared across different tasks to extract general features. $f_\phi$ and $f_{\gamma}$ are two amortized networks to estimate the posterior and prior distributions over $\omega$. $q$ and $p$ refer to posterior and prior distributions. $r_x$ are features extracted over samples drawn from $D \backslash C$. These features are l2-normalized as well as average pooled over samples in the batch. 

On the left, we show the posterior and the priors generated in the sequence of tasks. On the right, the inference model is depicted. To predict a label for a given query sample, first, the input images and its corresponding coreset are forwarded through $h_\theta$ and their features are computed: $r_x^t$ and $r_c^t$. Next, we feed $r_c^t$ through $f_\phi$ and estimate the posterior distribution over $q_\phi(\omega \mid \mathcal{C}_t)$. Then, we create the random Fourier bases $\omega_t$ by drawing samples from estimated posterior distribution. Having random bases for the current task $t$, $\omega_t$, as well as $r_x^t$ and $r_c^t$, random Fourier features related the query input $\psi(r_x^t)$ and coreset data $\psi(r_c^t)$ are estimated. Each kernel, $K$ and $\tilde{K}$, is estimated using its corresponding random Fourier features and $\mathrm{k}\left(\mathbf{x}, \mathbf{x}^{\prime}\right) {=} \psi(\x)\psi(\x^\prime)^\top$. Based on Eq. (5) in the main manuscript, these two estimated kernels are used to predict the output labels for given query samples.

Note that for the variant of our variational random features with an uninformative prior. the prior network is removed and the prior is set to a standard Gaussian distribution. In addition, by using linear, polynomial, and radial basis function kernels, neither the prior network nor the posterior network is used.

\subsection{Hyperparameters}
\label{sec:hparams}
We provide the detailed hyperparameter settings in Table~\ref{tab:hparams}, which are used to to generate Figure 7 and Table 3 in the main paper for each dataset.

\begin{table*}[!htp]
\centering
\caption{Hyperparameters used to generate results in Figure 7 and Table 3.}
\vspace{2mm}
\resizebox{.8\textwidth}{!}{%
\begin{tabular}{lcccc}
\toprule
\textbf{Method} 
& \textbf{Permuted MNIST} & \textbf{Rotated MNIST} & \textbf{Split CIFAR100} & \textbf{Split miniImageNet} \\ \midrule
Batch Size       & 10   & 10   & 10   & 10   \\
Learning Rate (LR)    & 0.1  & 0.1  & 0.3 & 0.3   \\
LR Decay Factor     & 0.8  & 0.8  & 0.95  & 0.95   \\
Momentum         & 0.8  & 0.8  & 0.8  & 0.8   \\
Dropout          & 0.5  & 0.5  & 0.02  & 0.02  \\
Coreset Size     & 20   &   20 &   20 &   30  \\
Number of Bases  & 1024 & 1024 & 2048 & 2048  \\
Number of Tasks  & 20   & 20   & 20   &  20 \\
Tau              & 0.01 & 0.01 & 0.01 & 0.01  \\
\bottomrule
\end{tabular}
}
\vspace{-2mm}
\label{tab:hparams}
\end{table*}

% \section{Extra Discussion over Ablation Study}
% \label{sec:paper_ablation_discussion}
% In the paper, we provide different ablation studies in order to experiment how different parts of our proposed method work. 

\subsection{Additional ablation  results on miniImageNet}
\label{sec:paper_ablation_discussion}
We provide additional ablation results on the miniImageNet dataset. We report the influence of the inference memory and the number of Random bases.  Figure~\ref{fig:core_vs_acc} shows increasing the coreset size from $1$ to $20$, improves average accuracy consistently. It saturates between $20$ to $40$. By enlarging the coreset size to 50, model performance increases again. In Table 3 in the main paper the results related to miniImageNet are reported based on a coreset size of 30. Figure~\ref{fig:basis_vs_acc} highlights how the number of random bases affects the average accuracy. Consistent with the findings in other datasets, the performance increases with a larger number of random Fourier bases. Since miniImageNet is a challenging benchmark, we choose $2048$ as the number of bases for all remaining experiments in the main paper.

\begin{figure*}[!h]
\hfill
\minipage[t]{0.45\linewidth}
  \includegraphics[width=\linewidth]{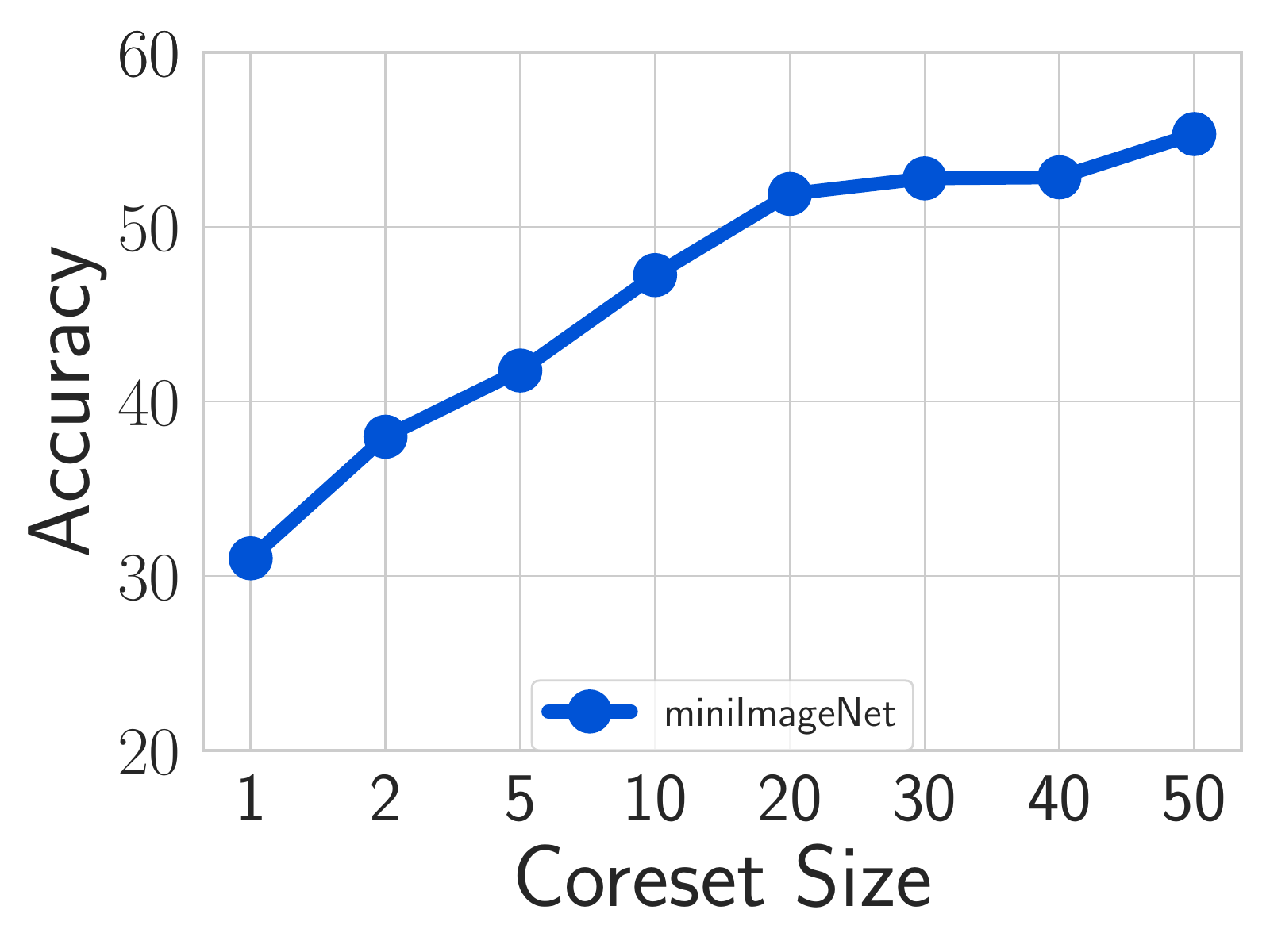}
  \caption{\textbf{How much inference memory?} Enlarging the coreset size of the VRF kernel leads to improvement of performance on miniImageNet benchmark dataset. Coreset size 30 is chosen for conducting experiment in the main paper.}
  %, we recommend a size of 30 for miniImageNet and 20 for the other three datasets. }
  \label{fig:core_vs_acc}
\endminipage\hfill
\minipage[t]{0.45\linewidth}%
  \includegraphics[width=\linewidth]{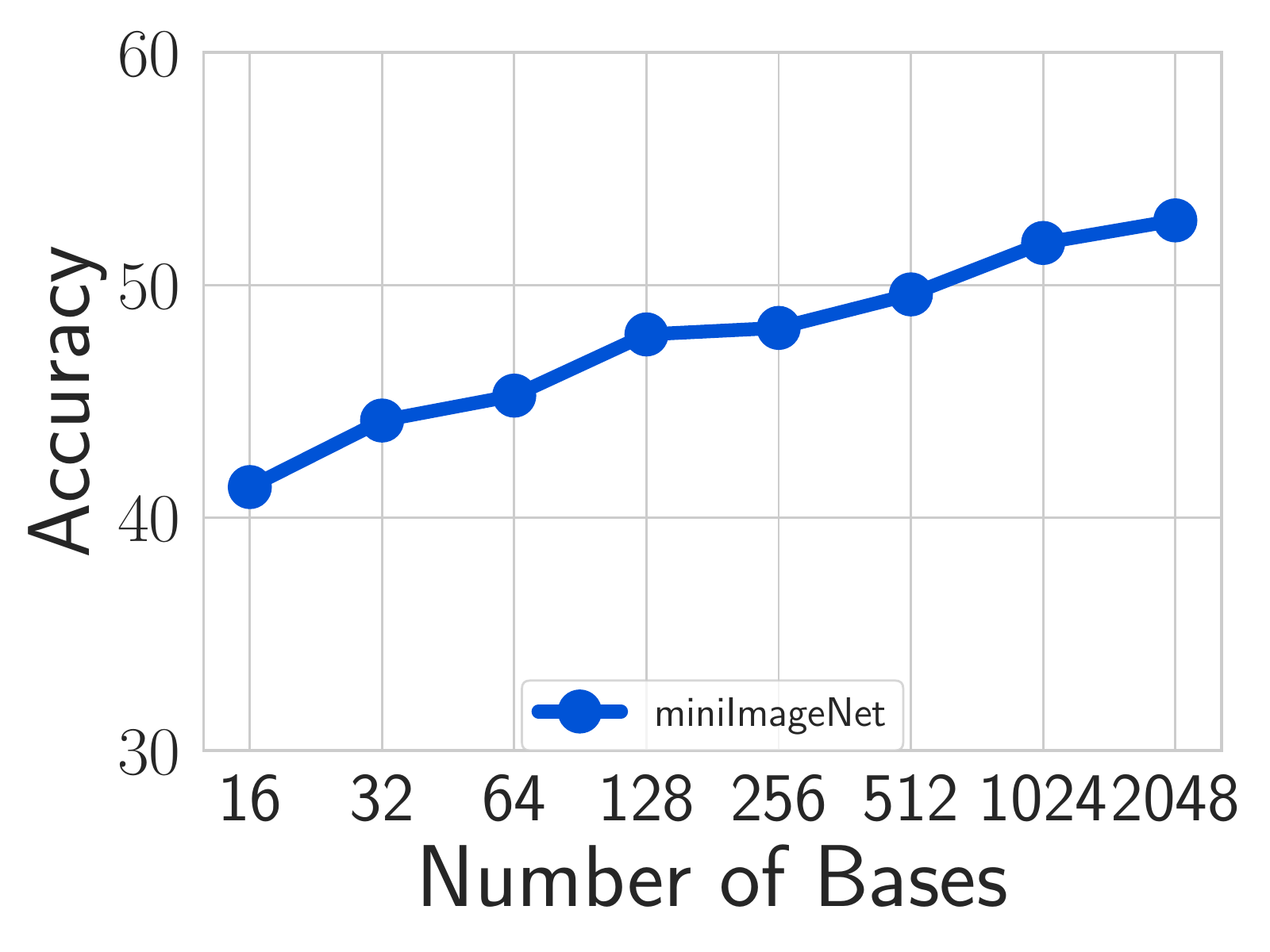}
  \caption{\textbf{How many Random Bases?} In general, a larger number of random Fourier bases consistently improves  performance on miniImageNet benchmark dataset. In the main paper, number of bases is set to be 2048.
  }
  \label{fig:basis_vs_acc}
\endminipage
\end{figure*}

\end{document}